%% file: main.tex
\documentclass[runningheads]{llncs}

\usepackage{eccv}

\usepackage{colortbl} 
\usepackage{booktabs}
\usepackage{amsmath}
\usepackage{amsfonts}
\usepackage{bm}         
\usepackage{multirow}   
\usepackage{tabularx}   
\usepackage{colortbl}   
\usepackage{booktabs}   

\usepackage{pgfgantt}
\usepackage{tikz}
\usetikzlibrary{arrows.meta,decorations.pathreplacing,calligraphy}

\definecolor{mygreen}{RGB}{204,255,204} 
\definecolor{myyellow}{RGB}{245,245,190}
\definecolor{myred}{RGB}{255,204,204} 

\usepackage{color,soul} 

\usepackage{eccvabbrv}

\usepackage{graphicx}
\usepackage{booktabs}

\usepackage[accsupp]{axessibility}  

\usepackage[pagebackref]{hyperref}

\usepackage{orcidlink}

\usepackage{makecell}
\usepackage[export]{adjustbox}

\usepackage{acro}
\DeclareAcronym{ML}{
  short = ML,
  long  = Machine Learning,
  sort  = ML,
}

\DeclareAcronym{MU}{
  short = MU,
  long  = Machine Unlearning,
  sort  = MU,
}

\DeclareAcronym{AI}{
  short = AI,
  long  = Artificial Intelligence,
  sort  = AI,
}

\DeclareAcronym{CNN}{
  short = CNN,
  long  = Convolutional Neural Network,
  sort  = CNN,
  plural = s,
}

\DeclareAcronym{GAN}{
  short = GAN,
  long  = Generative Adversarial Network,
  plural = s,
}

\DeclareAcronym{cGAN}{
  short = cGAN,
  long  = Conditional Generative Adversarial Network,
  plural = s,
}

\DeclareAcronym{GDPR}{
  short = GDPR,
  long  = General Data Protection Regulation,
}

\DeclareAcronym{FIM}{
  short = FIM,
  long  = Fisher Information Matrix,
}

\DeclareAcronym{EWC}{
  short = EWC,
  long  = Elastic Weight Consolidation,
}

\DeclareAcronym{PEFT}{
  short = PEFT,
  long  = Parameter Efficient Fine-Tuning,
}

\DeclareAcronym{PEM}{
  short = PEM,
  long  = Parameter Efficient Module,
}

\DeclareAcronym{OMP}{
  short = OMP,
  long  = One-shot Magnitude Pruning,
}

\DeclareAcronym{SPARE}{
  short = SPARE,
  long  = Self-distillation for PARameter Efficient Removal,
}

\DeclareAcronym{SFD}{
  short = SFD,
  long  = Score Forgetting Distillation,
}

\DeclareAcronym{DDPM}{
  short = DDPM,
  long  = Denoising Diffusion Probabilistic Models,
}

\DeclareAcronym{FID}{
  short = FID,
  long  = Fréchet Inception Distance,
}

\begin{document}

\title{SPARE: Self-distillation for PARameter-Efficient Removal} 



\author{
    Natnael Mola\inst{1}
    \and Leonardo S. B. Pereira\inst{1} \orcidlink{0000-0001-9429-7308}
    \and Carolina R. Kelsch\inst{1}
    \and Luis H. Arribas\inst{2}
    \and Juan C. S. M. Avedillo\inst{1}
}

\authorrunning{N.~Mola et al.}

\institute{Universidad Autonoma de Madrid, Spain
\and Universidad Politécnica de Madrid, Spain}

\maketitle

\begin{abstract}
  Machine Unlearning aims to remove the influence of specific data or concepts from trained models while preserving overall performance, a capability increasingly required by data protection regulations and responsible AI practices. Despite recent progress, unlearning in text-to-image diffusion models remains challenging due to high computational costs and the difficulty of balancing effective forgetting with retention of unrelated concepts. We introduce \ac{SPARE}, a two-stage unlearning method for image generation that combines parameter localization with self-distillation. SPARE first identifies parameters most responsible for generation of the unwanted concepts using gradient-based saliency and constrains updates through sparse low rank adapters, ensuring lightweight, localized modifications. In a second stage, SPARE applies a self-distillation objective that overwrites the unwanted concept with a user-defined surrogate while preserving behavior for other concepts. In addition we proposed a timestep sampling scheme for diffusion models to target only the crucial timesteps for a given concept leading to efficient unlearning. SPARE surpasses the current state-of-the-art on the UnlearnCanvas benchmark, and ablation studies on several datasets indicate fine-grained control over the forgetting-retention trade-off. Our results demonstrate that SPARE achieves strong concept erasure and high retainability across various domains, making it a suitable solution for selective unlearning in diffusion-based image generation models.
\end{abstract}

\section{Introduction}
\label{sec:intro}

\ac{MU} is the area of \ac{ML} that studies the removal of data from an \ac{AI} system, making the system unlearn a concept or forget what it learned from a specific data point \cite{shaik2024}.
This area has been given more importance over the past few years with the advancement of some regulations that provide users with the right to have more control over the data they generate, such as the \ac{GDPR} in the EU \cite{gdpr2016}. With that in force, companies need to be able to remove from their trained models data from specific users. Besides that, \ac{MU} can be used for mitigating biases \cite{zhang2023} and improving model interpretability \cite{choi2025mu_interpretation}, thus contributing to greater efforts towards safe and responsible AI.

The field has evolved in a modality-siloed manner, where most methods are evaluated either on textual tasks, or on visual classification, or on text-to-image generation, but rarely across modalities \cite{xu2023review}. In such a fragmented scenario, image generation tasks have lagged behind in the pace of development, partly because diffusion models require substantially higher compute to fine-tune \cite{ho2020denoising}, and partly because evaluating unlearning in generation is intrinsically harder \cite{fan2023salun}.

Existing unlearning methods for image generation struggle to balance the removal of the target concept with the preservation of unrelated capabilities \cite{shaik2024, Cai2025Slug}.  
This balance is essential: aggressive updates often trigger collateral forgetting, where semantically or statistically adjacent concepts degrade together, reducing the practical utility of the model.  
Furthermore, most current approaches provide limited mechanisms to control the trade-off between forgetting and retention \cite{wang2024selectiveforgetting}, which restricts their applicability in real-world settings that require selective and explainable unlearning at scale.

In this work, we present a new \ac{MU} method for image generation named Self-distillation for PARameter-Efficient Removal (SPARE), a two-stage unlearning method that first identifies the subset of parameters most responsible for the generation of the unwanted concept through a per-weight saliency mask and then applies a lightweight overwrite procedure using self-distillation based objective. The update is implemented through sparse LoRA adapters, which restrict all modifications to the selected weights and keep the memory footprint low, moreover, the adapter can be merged or removed at runtime, enabling reversible deployment. We also made experimental analysis showing different concepts have different crucial timesteps during inference and we proposed a custom timestep sampling scheme during training to target these timesteps for efficient unlearning. SPARE allows the user to move along the Pareto front of solutions, deciding how aggressively forgetting should be performed and at what expense of retention.
The entire implementation is integrated into the 
Vision-Unlearning open-source library
\footnote{
https://pypi.org/project/vision-unlearning/
}.

We evaluate SPARE on the UnlearnCanvas benchmark \cite{Zhang2024UnlearnCanvas}, alongside small-scale ablation studies on the dataset Imagenette \cite{imagenette}, Labeled Faces in the Wild (LFW) \cite{dataset_lfw}, AtharvaTaras Dog Breeds Dataset
\cite{dataset_taras_dog_breeds}, and Scene UNderstanding (SUN) Attributes \cite{dataset_sun_attributes}. SPARE improves significantly over the State-of-the-Art while allowing an unprecedented amount of control over how forgetting and retaining are balanced. Furthermore, our method results in lightweight modules that can be added and removed from a model at runtime, respectively forgetting or restoring the original behavior, easing its usage in production environments.

\section{Literature Review}

\subsection{Diffusion Models}
Diffusion models \cite{ho2020denoising} are latent variable models that generate data by reversing a stochastic noising process. The fixed forward process progressively corrupts a data sample $\mathbf{x}_0 \sim q(\mathbf{x}_0)$ into isotropic Gaussian noise over $T$ steps. Using a variance schedule $\beta_t$, where $\alpha_t = 1 - \beta_t$ and $\bar{\alpha}_t = \prod_{s=1}^t \alpha_s$, any intermediate state $\mathbf{x}_t$ can be sampled in closed form:
\begin{equation}
    q(\mathbf{x}_t | \mathbf{x}_0) = \mathcal{N}(\mathbf{x}_t; \sqrt{\bar{\alpha}_t} \mathbf{x}_0, (1 - \bar{\alpha}_t) \mathbf{I})
\end{equation}

To generate data, the model learns the reverse process to recover $\mathbf{x}_0$ from pure noise $\mathbf{x}_T \sim \mathcal{N}(\mathbf{0}, \mathbf{I})$. Specifically, a neural network $\boldsymbol{\epsilon}_\theta(\mathbf{x}_t, t)$ is trained to predict the injected noise $\boldsymbol{\epsilon} \sim \mathcal{N}(\mathbf{0}, \mathbf{I})$, optimized via a simplified variational lower bound:
\begin{equation}
    \mathcal{L}_{\text{simple}} = \mathbb{E}_{t, \mathbf{x}_0, \boldsymbol{\epsilon}} \left[ \| \boldsymbol{\epsilon} - \boldsymbol{\epsilon}_\theta(\mathbf{x}_t, t) \|_2^2 \right]
\end{equation}

Stable Diffusion \cite{rombach2022stableDiffusion} optimizes this framework by operating in a compressed latent space rather than pixel space. By coupling a UNet backbone for denoising, a variational autoencoder (VAE) for latent compression, and a frozen CLIP encoder for text conditioning, it enables high-fidelity generation at a highly manageable computational cost.

\subsection{Parameter-Efficient Fine-Tuning (PEFT)}

Parameter-Efficient Fine-Tuning (PEFT) refers to techniques that adapt large pre-trained models to downstream tasks while updating only a small fraction of parameters, thus significantly reducing computation, memory, and storage costs relative to full fine-tuning \cite{xu2023}.  
A prominent instantiation is LoRA (Low-Rank Adaptation), which freezes the original model weights and injects trainable low-rank matrices into selected layers, allowing the adaptation to be expressed as the product of two small matrices rather than modifying the full weight matrix \cite{hu2021lora}.  

Beyond standard LoRA, the work of \cite{munoz2024_sparse_peft} introduces SparseLoRA, a variation designed to constrain adaptation so that only a small, deliberately selected subset of weights in the frozen model is modified with the goal of adapting a sparse model without affecting the initial sparsity. This is achieved by using a curated binary mask to make the update only for the selected parameters.


\subsection{Machine Unlearning}


The unlearning task can be formalized as the modification of a \ac{ML} model in order to remove the influence and the knowledge learned from part of the dataset, denoted as \textit{forget set} $(D_f)$, while retaining knowledge obtained from training in the remaining data, denoted as \textit{retain set} $(D_r)$ \cite{cao2015}. Optionally, some method \cite{kumari2023} use a set specifying the concept to which prompts referring to $D_f$ should map to, henceforth referred to as \textit{overwrite set} $(D_o)$.

\subsubsection{Information-theoretic approaches}
Information-theoretic approaches to machine unlearning (MU) frequently leverage the Fisher Information Matrix (FIM) to quantify the amount of information model parameters retain about specific data samples \cite{clavera2024}. For a distribution parameterized by $\theta$, the FIM elements are defined as $F_{ij}(\theta) = \mathbb{E} \left[ \frac{\partial \log p(x|\theta)}{\partial \theta_i} \frac{\partial \log p(x|\theta)}{\partial \theta_j} \right]$. Because computing and storing the exact FIM scales quadratically with the number of parameters, making it intractable for modern deep learning models, practical implementations rely on approximations such as the diagonal \cite{clavera2024} or Kronecker-factored \cite{martens2015kronecker} forms. These computationally efficient approximations enable techniques like Fisher Forgetting \cite{golatkar2020}, which achieves targeted unlearning by injecting Gaussian noise scaled by the forget set's FIM into the weights, and Influence Unlearning \cite{izzo2020}, which utilizes the FIM to approximate the inverse Hessian for a projective residual update, successfully erasing specific data influences without full retraining.

\subsubsection{Sparse-update approaches} \label{literature_sparsity}
Weight sparsity, first formalized in the Lottery Ticket Hypothesis \cite{Frankle2019}, has been repeatedly adopted in \ac{MU} as a mechanism to constrain parameter updates and limit collateral forgetting.  
In this setting, sparsity is treated as a saliency or ``location'' map \cite{Lee2025Location}: the unlearning algorithm updates only a selected subset of parameters, ideally those most responsible for representing the forget set, while freezing the rest of the network.  
The rationale is simple: restricting the update space preserves unrelated concepts and helps maintain global model behavior.  
Despite legitimate criticism about the stability and interpretability of saliency-based selection \cite{Lee2025Location}, several works report measurable reductions in unintended forgetting when such mechanisms are used \cite{fan2023salun, Cai2025Slug}.

\subsubsection{Distillation-based approaches}
Distillation-based approaches to machine unlearning adapt knowledge distillation \cite{Hinton2015self_distillation} to steer a student model to forget specific concepts while maintaining global generation quality by constraining it to match a frozen teacher model on retained data \cite{chen2025scoreforgettingdistillation}. Various methods implement this decoupling of forgetting and preservation, such as Concept Ablation (CA) \cite{kumari2023}, which explicitly projects out concept-specific directions from cross-attention activations, and frameworks like UNDO \cite{Lee2025UNDO}, PURGE \cite{Quan2025Purge}, and DELETE \cite{Zhou2025Delete}, which utilize decoupled, partitioned, or logits-level distillation targets. In the context of diffusion models, Score Forgetting Distillation (SFD) \cite{chen2025scoreforgettingdistillation} formalizes an explicit \textit{overwrite} mechanism (also known as alignment \cite{Gandikota2023UCE} or mapping \cite{George2025}). Instead of merely erasing a concept, the student is trained to match the teacher's noise predictions conditioned on a semantically distant replacement prompt (\eg, replacing ``Brad Pitt'' with ``middle-aged man''). While this overwrite mechanism successfully prevents the re-emergence of unlearned concepts and can be integrated beyond distillation, it complicates evaluation. Because standard metrics \cite{heusel2018fidmetric, Radford2021Clip} are highly sensitive to the semantic distance of the chosen replacement, quantitative comparisons necessitate explicit documentation of the overwrite choice to ensure fair evaluation \cite{George2025}.

\section{SPARE}

SPARE  has two stage approach for unlearning which involves knowledge localization in the first step to identify weights that are the most important for the current forgetting task, and in a second moment it updates the model via distillation based objective. \cref{fig_fade_schematic} shows the overview of our SPARE mechanism.

\subsection{Knowledge location step}



To mitigate catastrophic forgetting and preserve the model's general capabilities, we restrict parameter updates strictly to the weights responsible for the targeted concept. We achieve this by evaluating the saliency of the pre-trained weights $W$ with respect to the forget set $D_f$, isolating a subset of weights $W_f$ whose saliency scores exceed a predefined threshold $\gamma$, inspired in SalUn \cite{fan2023salun}.  

To guarantee that the LoRA modules exclusively alter the specific parameters within $W_f$, we introduce a masking mechanism inspired by SparsePEFT \cite{munoz2024_sparse_peft}. We construct a binary mask $M$ that acts as an indicator function for the salient weights. During training, the low-rank update, which is defined by the product of the low rank adapter matrices $B$ and $A$, is constrained via a Hadamard product ($\odot$) with this mask. For an input $x$, the modified forward pass is formulated as:
\begin{equation}
    h = Wx + ((BA) \odot M)x, \quad \text{where} \quad M = \mathbb{I}(\mathcal{S}(W, D_f) > \gamma)
\end{equation}
Here, $\mathcal{S}(\cdot)$ denotes the saliency scoring function (the gradient of the the loss when computed using samples in $D_f$ for all parameters in $W$) and $\mathbb{I}(\cdot)$ represents the indicator function, assigning a value of $1$ for the parameters that satisfy the condition and $0$ for the rest of the parameters.

While this formulation applies a fine-grained, per-weight mask, we also explore a coarse-grained, per-block masking alternative. We systematically compare the performance of both the fine-grained and coarse-grained approaches in our ablation study and contextualize them within the broader literature on sparse adaptation in \cref{literature_sparsity}. The source code for this specific SparsePEFT variant has been released independently to facilitate further research 
\cite{sparse_peft}
.

\subsection{Distillation training step}

The second phase of our methodology aims to selectively eradicate knowledge associated with $D_f$ from the salient weights $W_f$ via a self-distillation objective, drawing inspiration from Score Forgetting Distillation \cite{chen2025scoreforgettingdistillation} and Concept Ablation \cite{kumari2023}. To preserve general capabilities and generation quality on non-target domains, we enforce alignment between the noise predictions of the actively fine-tuned student model $\boldsymbol{\epsilon}_\theta$ and a frozen reference teacher $\boldsymbol{\epsilon}_{\theta_{\text{ref}}}$. Specifically, for a given image and prompt pair $(\mathbf{x}, \mathbf{c}_r)$ from the retain set $D_r$, and a diffused state $\mathbf{x}_t$ at a uniformly sampled timestep $t$, the models are aligned using the standard score matching objective:
\begin{equation}
    \mathcal{L}_{\text{retain}} = \mathbb{E}_{\mathbf{x}, \mathbf{c}_r \sim D_r, t, \boldsymbol{\epsilon}} \left[ \| \boldsymbol{\epsilon}_{\theta_{\text{ref}}}(\mathbf{x}_t, \mathbf{c}_r, t) - \boldsymbol{\epsilon}_\theta(\mathbf{x}_t, \mathbf{c}_r, t) \|_2^2 \right]
\end{equation}

Conversely, for samples within the forget set $(\mathbf{x}, \mathbf{c}_f) \in D_f$, we employ an explicit concept overwrite mechanism. The student model, conditioned on the target forget prompt $\mathbf{c}_f$, is optimized to match the noise predictions of the frozen teacher conditioned on a predefined, safe override concept $\mathbf{c}_o$:
\begin{equation}
    \mathcal{L}_{\text{forget}} = \mathbb{E}_{\mathbf{x}, \mathbf{c}_f \sim D_f, t, \boldsymbol{\epsilon}} \left[ \| \boldsymbol{\epsilon}_{\theta_{\text{ref}}}(\mathbf{x}_t, \mathbf{c}_o, t) - \boldsymbol{\epsilon}_\theta(\mathbf{x}_t, \mathbf{c}_f, t) \|_2^2 \right]
\end{equation}

The overall training objective is formulated as a weighted sum of the retention and forgetting distillation losses:
\begin{equation}
    \mathcal{L}_{\text{total}} = \lambda_{\text{retain}} \mathcal{L}_{\text{retain}} + \lambda_{\text{forget}} \mathcal{L}_{\text{forget}}
\end{equation}
where the hyperparameters $\lambda_{\text{retain}}$ and $\lambda_{\text{forget}}$ control the trade-off between preserving the original model's capabilities and enforcing the target erasure. Crucially, while this objective evaluates the full output of the network, the gradients derived from $\mathcal{L}_{\text{total}}$ are backpropagated exclusively to update the parameters of the sparse low-rank adapter. Because the update is constrained by the saliency mask $M$, the loss strictly optimizes the localized subset of weights $W_f$, leaving the vast majority of the pre-trained network entirely frozen. This isolated optimization strategy ensures that the unlearning process is both highly targeted and computationally efficient.

In order to leverage the sparse adapter obtained at the first step, we adapted the self-distillation for using LoRA. Furthermore, this method does not require maintaining a copy of the frozen model on the GPU, decreasing the VRAM usage, because it requires only enabling and disabling the LoRA module when needed: when we need the output of the original model, we disable the adapter and make the forward pass; when we want the output of the fine-tuned model, we enable the adapter.

\begin{figure}
    \centering\includegraphics[width=0.6\linewidth]{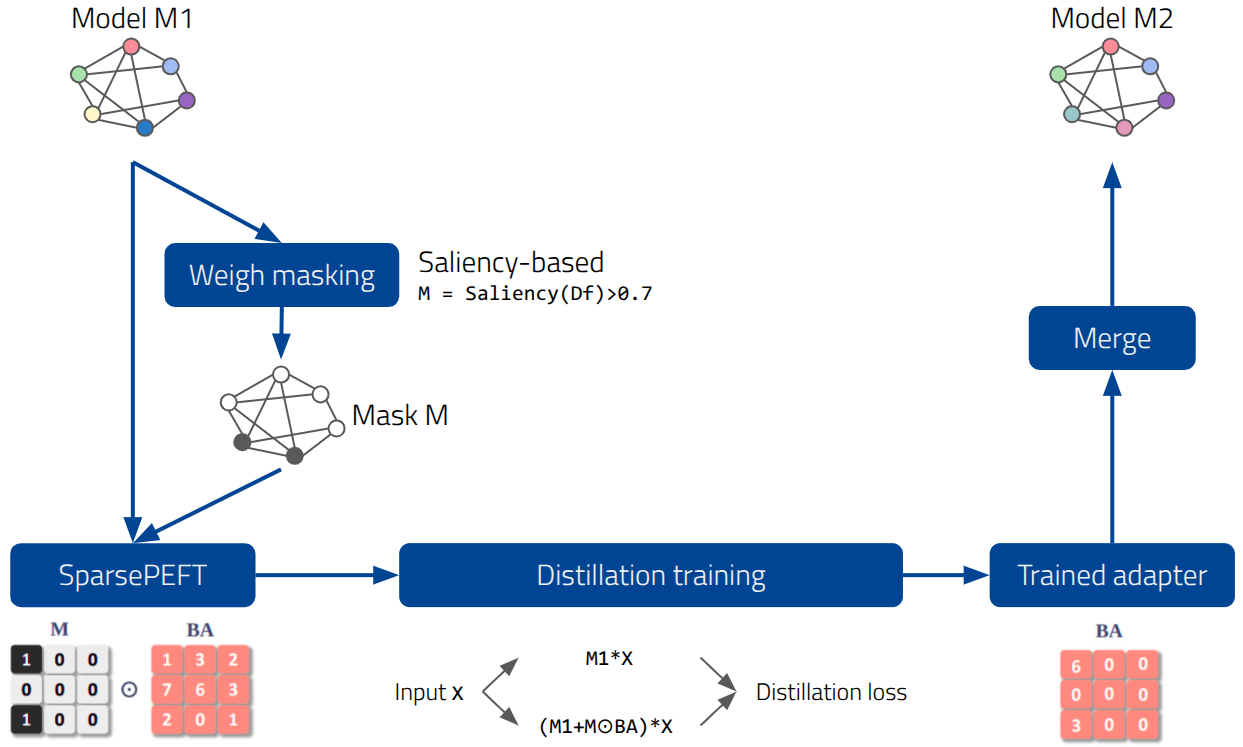}
    \caption{Schematic summary of our method. As shown on the left, binary mask will be computed with gradient based saliency and it will be used to make the product of the LoRA matrices sparse accordingly. The fine-tuning is done with self-distillation by toggling the LoRAs on and off as shown at the bottom. After fine-tuning, the LoRAs will be merged to the model for inference.}
    \label{fig_fade_schematic}
\end{figure}

Compared to the distillation scheme of Score Forgetting Distillation \cite{chen2025scoreforgettingdistillation}, our method calculates the loss directly on the image space, without using an auxiliary score network, since both teacher and student are computed for the same number of forward diffusion process steps.
Compared to Concept Ablation \cite{kumari2023}, our method add a preservation term, similar to the idea proposed in Memory Replay GANs \cite{wu2018_memory_replay_gans}, thus improving the retention quality; Furthermore, our method does not rely on stopgrad to obtain the
teacher's prediction. After the training is completed, the trained adapter can be safely merged into the original model (\cref{fig_fade_schematic}). This merge can be performed at runtime, and undone at any time, easing usage in production environments. 

\subsection{Timestep Targeting}

\begin{figure}[t]
    \centering
    \includegraphics[width=0.8\linewidth]{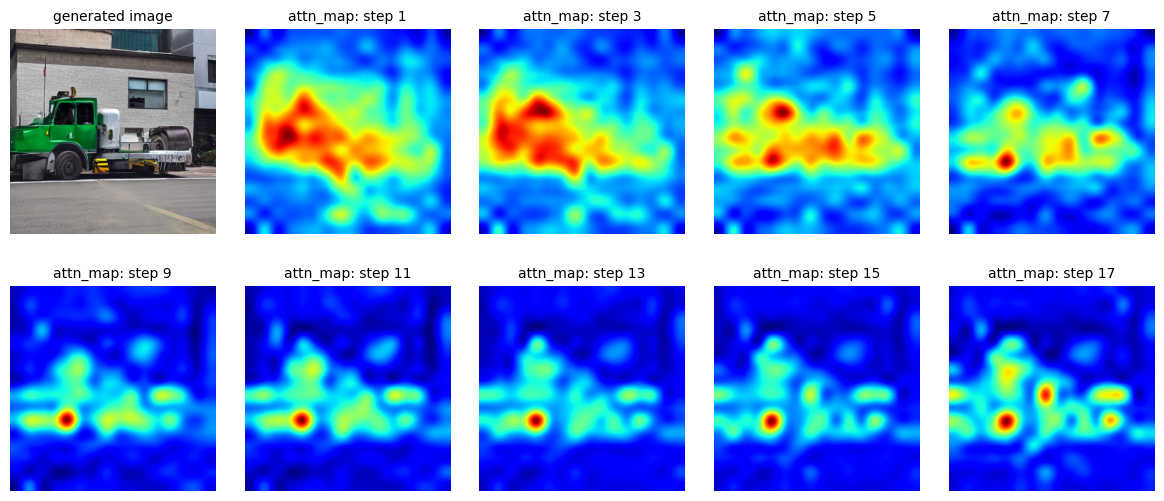}
    \caption{Progression of cross-attention maps between the token ``truck'' and the spatial image patches across diffusion timesteps, conditioned on the prompt ``an image of a truck'' during a 20 step inference, which demonstrates cross-attention maps becoming sparser}
    \label{fig:attn_ft_timesteps}
\end{figure}

Diffusion models generate images by iteratively denoising a sample drawn from a Gaussian prior over a predefined number of timesteps. Rather than contributing equally to the final output, extensive research has demonstrated that different timesteps serve distinct functional purposes during this progressive generation \cite{balaji2022ediff, hertz2022prompttoprompt}. Specifically, the synthesis follows a coarse-to-fine trajectory: early timesteps (high noise levels) primarily govern the global structure, layout, and semantic alignment of the image, whereas later timesteps (low noise levels) are dedicated strictly to refining local textures and high-frequency details. We can visualize this behavior by examining the cross-attention maps between the text prompt tokens and the patches of the generated image. As depicted in \cref{fig:attn_ft_timesteps}, for a 20 timestep DDIM \cite{song2020ddim} inference, these cross-attention maps become progressively sparser as denoising advances, indicating that the text tokens exert highly localized fine-grained influence during the later stages of synthesis.

Building upon this observation, we propose a targeted unlearning strategy that exclusively optimizes the specific timesteps crucial for a given concept, thereby improving unlearning efficiency. Since concepts vary in their defining characteristics, some relying heavily on global shape, others on color or texture, the optimal intervention window varies accordingly. In the context of concept overwrite mechanisms, unlearning can be localized to the specific timesteps where the generative trajectories of the original and replacement concepts diverge. 

\begin{figure}[htbp]
    \centering
    \begin{subfigure}[b]{0.7\textwidth}
        \centering
        \includegraphics[width=\textwidth]{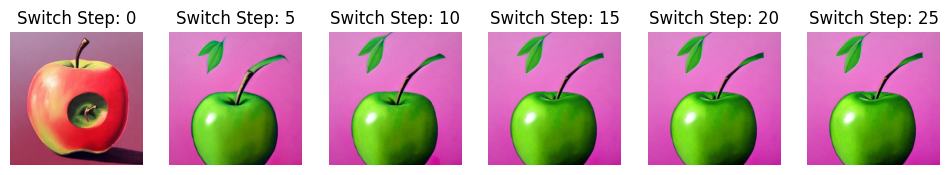}
        \caption{Prompt-switching inference from ``green apple'' to ``red apple'' at varying timesteps ($T=30$ DDIM\cite{song2020ddim} steps).}
        \label{fig:top_img}
    \end{subfigure}
    
    \vspace{0.2cm} 
    
    \begin{subfigure}[b]{0.7\textwidth}
        \centering
        \includegraphics[width=\textwidth]{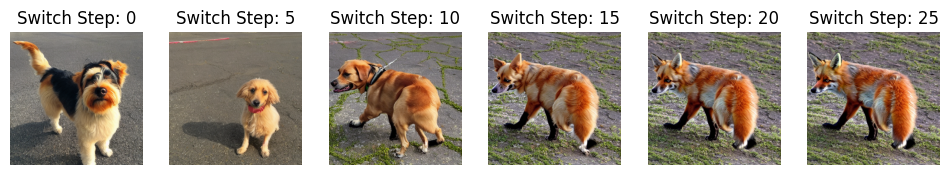}
        \caption{Prompt-switching inference from ``fox'' to ``dog'' at varying timesteps ($T=30$).}
        \label{fig:bottom_img}
    \end{subfigure}
    
    \caption{Demonstration of concept-dependent crucial timesteps via mid-generation prompt switching. \textbf{(a)} The semantic difference between green and red apples is resolved entirely in the earliest timesteps; switching the prompt from ``green apple'' to ``red apple'' after just 5 steps yields no change in color. \textbf{(b)} The structural divergence between a dog and a fox occurs later in the diffusion process, allowing the model to generate a dog even when conditioned on a fox for the first half of the timesteps.(Note in these figures, 0 indicates the first denoising timestep, and the images generated are generated purely by the second prompt)}
    \label{fig:stacked_subfigures}
\end{figure}

To empirically identify these critical timesteps, we developed a prompt-switching inference pipeline for Stable Diffusion, wherein the conditioning text prompt is swapped mid-generation. \cref{fig:stacked_subfigures} demonstrates that the point of semantic divergence is highly concept-dependent. For instance, the visual distinction between a ``green apple'' and a ``red apple'' is solidified during the very early stages of denoising; switching the prompt to ``red apple'' from ``green apple'' after these initial steps fails to alter the generated color. Conversely, the structural differences between a ``dog'' and a ``fox'' are resolved much later, allowing the generation to successfully pivot to a dog even if the initial trajectory was conditioned on a fox for the first half of the process. This confirms that the crucial timesteps for concept replacement are fundamentally concept-dependent, and that targeting these specific intervals yields a vastly more efficient and focused unlearning process.

\section{Experiments and Evaluation}\label{eval}
Here we present a series of evaluations of the proposed method in different datasets, a quantitative evaluation with an text-to-image generation benchmark, as well as some small scale ablation studies.

\subsection{Quantitative results}\label{quantitative-results}

\input{quantitative_eval}

\subsection{Custom Timestep Sampling results}

In standard diffusion model training, the timestep $t$ used to define the forward diffusion process is typically sampled uniformly from the entire defined trajectory, i.e., $t \sim \mathcal{U}(1, T)$, where $T = 1000$ for architectures like Stable Diffusion. Once $t$ is sampled, the original data $\mathbf{x}_0$ is corrupted to an intermediate noisy state with the forward diffusion.

However, as established in the preceding analysis, the generative trajectories of a target concept and its corresponding overriding concept diverge at specific, localized intervals rather than across the entire diffusion process. Consequently, we propose a targeted timestep sampling strategy that restricts the unlearning optimization strictly to these crucial timesteps. By focusing updates only where the concepts conflict, we can accelerate the unlearning process and minimize unintended degradation of the model's broader knowledge base. 

To empirically validate the impact of constrained timestep sampling on unlearning efficiency, we conducted a preliminary experiment focusing on style replacement using the UnlearnCanvas dataset. We evaluated the unlearning performance across three distinct sampling strategies for the forward diffusion process:

\begin{itemize}
    \item \textbf{Uniform sampling} (Baseline): $t \sim \mathcal{U}(1, 1000)$
    \item \textbf{Early-stage sampling} (High noise, structural layout): $t \sim \mathcal{U}(400, 1000)$
    \item \textbf{Late-stage sampling} (Low noise, stylistic refinement): $t \sim \mathcal{U}(1, 600)$
\end{itemize}

Further more, we proposed a custom sampling range based on the proposed prompt switching mechanism. To find a lower bound (note that low timesteps corresponds to late denoising steps) we find a timestep after which switching out the prompt will not change the generated concept. And for higher bound, we find a timestep before which if we use a different prompt will not change the results. 

\cref{tab:checkpoint50_shaded} shows the results for various style replacement finetuning we did with SPARE with the four timestep sampling strategies just for 50 unlearning steps. The metric used is the amount of time image generated by a model that unlearned a target style is classified as that target style. From the table, we can clearly observe the sampling scheme affects the unlearning accuracy and there is a pattern suggesting different styles have different crucial timesteps. For instance whenever the early sampling is better than the uniform, the late sampling performs the worst which is clear indicator of early timesteps being crucial for that specific style. Our proposed custom sampling scheme is the only one that performs better than uniform sampling on average. Also since uniform sampling covers every timesteps, it is possible for some cases it can sample the right timesteps by chance, hence performing better than the custom sampling in some cases.

\begin{table}[h!]
\centering
\caption{\textbf{Style unlearning accuracy across sampling schemes.} We compare various timestep windows for style unlearning ($A \to B$). \textbf{Bold} indicates the best overall performance ($\downarrow$); \underline{underline} indicates the second best. The \colorbox{gray!15}{shaded column} highlights our proposed Custom scheme, and an asterisk ($^*$) indicates where it strictly outperforms the Uniform baseline.}
\label{tab:checkpoint50_shaded}
\small
\begin{tabular}{l c c c >{\columncolor{gray!15}}c} 
\toprule
& \textbf{Late} & \textbf{Early} & \textbf{Uniform} & \textbf{Custom} \\ 
\textbf{Style Session ($A \to B$)} & (0--600) & (400--1000) & (0--1000) & \textbf{Range} \\ 
\midrule
Artist Sketch $\to$ Bricks     & 0.8200 & 0.5000 & \textbf{0.3800} & \underline{0.4200} \\
Bricks $\to$ Byzantine         & \textbf{0.0200} & 0.1600 & \textbf{0.0200} & \textbf{0.0200} \\
Cartoon $\to$ Cold Warm        & 1.0000 & \textbf{0.6600} & 0.8800 & \underline{0.7200}$^*$ \\
Color Fantasy $\to$ Comic Etch & \textbf{0.2600} & 0.7600 & 0.3400 & \underline{0.3200}$^*$ \\
Comic Etch $\to$ Crayon        & 0.6000 & 0.7200 & \textbf{0.3400} & \underline{0.4600} \\
Cubism $\to$ Dadaism           & \underline{0.4200} & 0.5600 & 0.4400 & \textbf{0.2000}$^*$ \\ 
Picasso $\to$ Cartoon          & \textbf{0.0000} & 0.5200 & 0.2200 & \underline{0.0400}$^*$ \\ 
\bottomrule
\end{tabular}
\end{table}

\cref{tab:clip_scores_celebs} compares the uniform and our custom sampling schemes for celebrity unlearning on the Labeled Faces in the Wild (LFW) dataset \cite{dataset_lfw}, using the CLIP score between the generated image and the target concept prompt. The results show that our custom sampling scheme consistently achieves better performance. Notably, the timestep ranges discovered by our prompt-switching mechanism align with visual intuition. For example, when replacing Angelina Jolie with Jennifer Lopez, or Colin Powell with Barack Obama, the method selects later timesteps. This makes sense, as these pairs share basic structural similarities. Conversely, when replacing Angelina Jolie with Vin Diesel, two identities with very different appearances, the method selects earlier timesteps, reflecting the need to alter the overall image structure.

\begin{table}[h!] 
\centering
\caption{\textbf{Quantitative comparison on Identity unlearning} CLIP scores ($\downarrow$) between the target concept prompt and generated images for uniform vs. our custom sampling scheme across different checkpoints (ckpts). $A \to B$ denotes the unlearning of concept $A$ towards target $B$. \textbf{Bold} indicates the best (lowest) score for each checkpoint.}
\label{tab:clip_scores_celebs}
\small 
\resizebox{\textwidth}{!}{
\begin{tabular}{lcccccc}
\toprule
& & \multicolumn{2}{c}{\textbf{ckpt-100}} & \multicolumn{2}{c}{\textbf{ckpt-200}} & \textbf{Custom} \\ 
\cmidrule(lr){3-4} \cmidrule(lr){5-6}
\textbf{Unlearning Session} & \textbf{Baseline} & \textbf{Uniform} & \textbf{Custom} & \textbf{Uniform} & \textbf{Custom} & \textbf{Range} \\ 
\midrule
Angelina Jolie $\to$ Vin Diesel & 0.3221 & 0.3118 & \textbf{0.3037} & 0.2883 & \textbf{0.2874} & $[300, 900]$ \\
Angelina Jolie $\to$ J.Lopez    & 0.3221 & 0.3048 & \textbf{0.3025} & 0.2912 & \textbf{0.2896} & $[100, 600]$ \\
G.W. Bush $\to$ M.Jackson       & 0.3186 & 0.2379 & \textbf{0.2335} & 0.2294 & \textbf{0.2243} & $[300, 850]$ \\
Colin Powell $\to$ Obama        & 0.3456 & 0.2731 & \textbf{0.2642} & 0.2630 & \textbf{0.2621} & $[100, 650]$ \\ 
\bottomrule
\end{tabular}}
\end{table}

A limitation of our custom sampling scheme is that it requires either human evaluation or trained models to determine which of the two prompts the generated image resembles. While pre-trained vision-language models like CLIP can often be used for zero-shot classification between the two prompts, this approach is not always reliable. We also experimented with measuring the similarity between the cross-attention maps of the two prompts. However, this method was ineffective; due to a strong center bias in many generated images, the cross-attention maps were too difficult to distinguish from one another clearly.

\begin{table}[h!]
\centering
\setlength{\tabcolsep}{2pt}          
\renewcommand{\arraystretch}{1}     
\setlength{\extrarowheight}{0pt}    

\begin{tabular}{c|c|c|
m{1.8cm}
m{1.8cm}
m{1.8cm}
m{1.8cm}
}
\hline
\multirow{2}{*}{Task} &
\multirow{2}{*}{\makecell{Forget \\Concept}} &
\multirow{2}{*}{\makecell{Overwrite \\Concept}} &
\multicolumn{2}{c}{Forget} &
\multicolumn{2}{c}{Retain} \\

& & &
\multicolumn{1}{c}{Unlearned} & \multicolumn{1}{c}{Original} &
\multicolumn{1}{c}{Unlearned} & \multicolumn{1}{c}{Original} \\
\hline

People &
\makecell{George \\W. Bush} &
Kid &
\includegraphics[width=\linewidth]{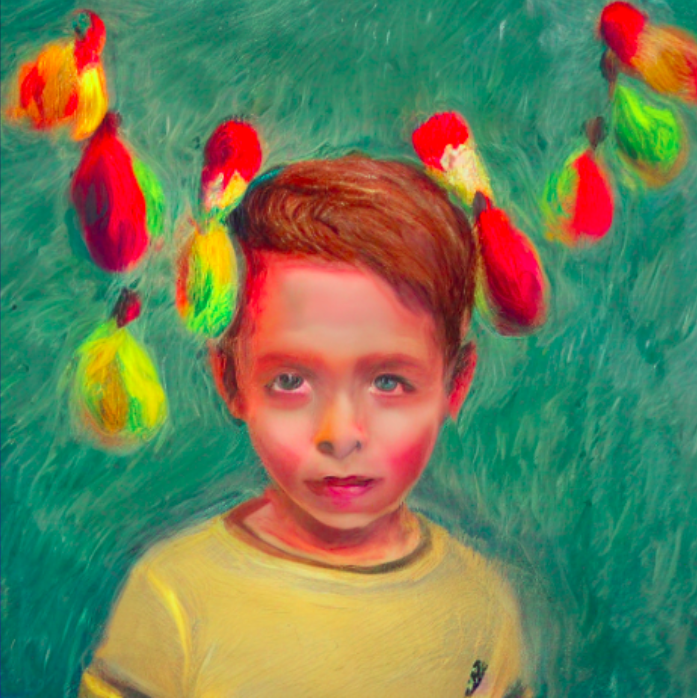} &
\includegraphics[width=\linewidth]{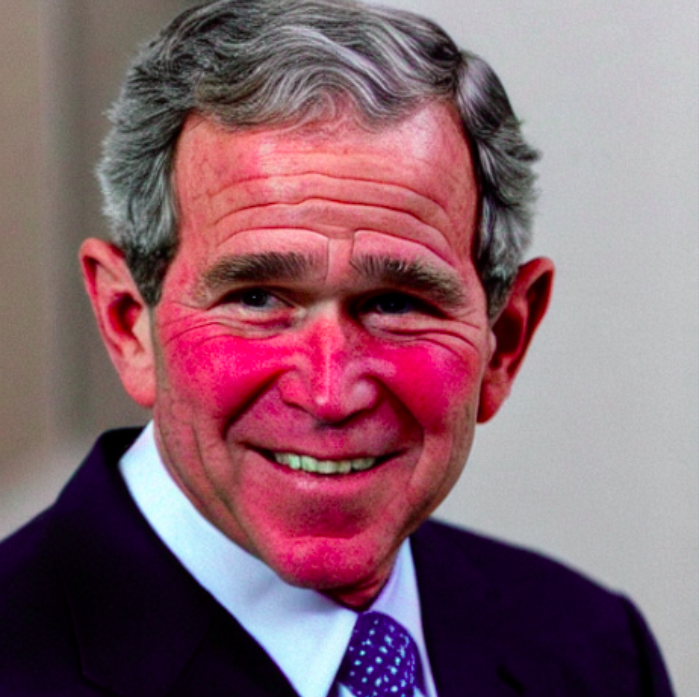} &
\includegraphics[width=\linewidth]{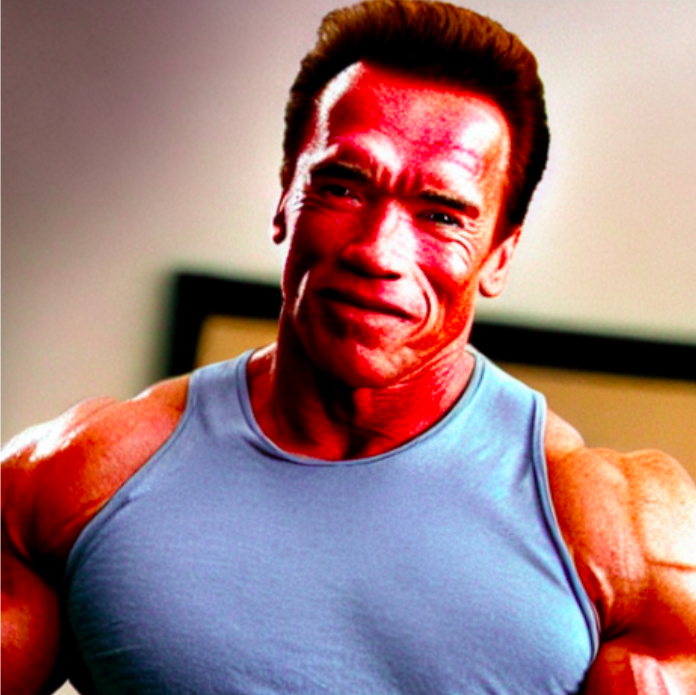} &
\includegraphics[width=\linewidth]{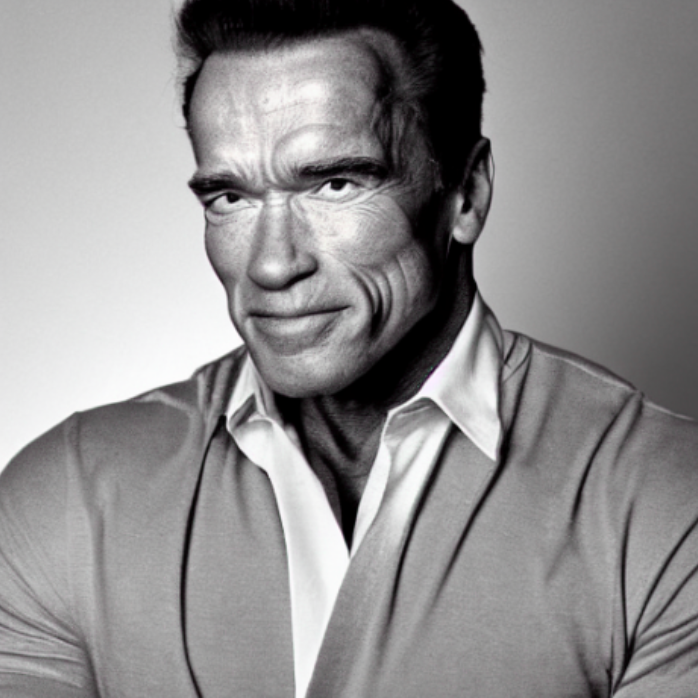} \\
\makecell{Dog \\breed} &
\makecell{Griffon \\Bruxellois} &
Cat &
\includegraphics[width=\linewidth]{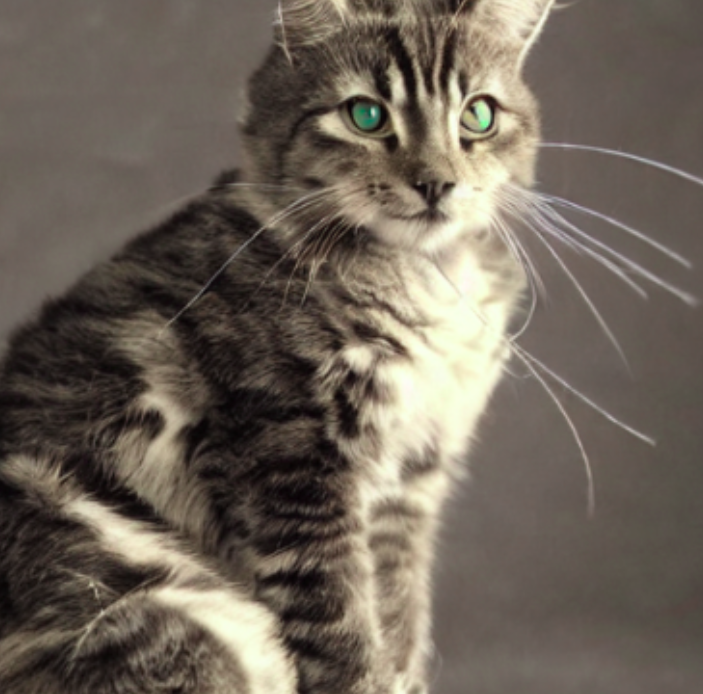} &
\includegraphics[width=\linewidth]{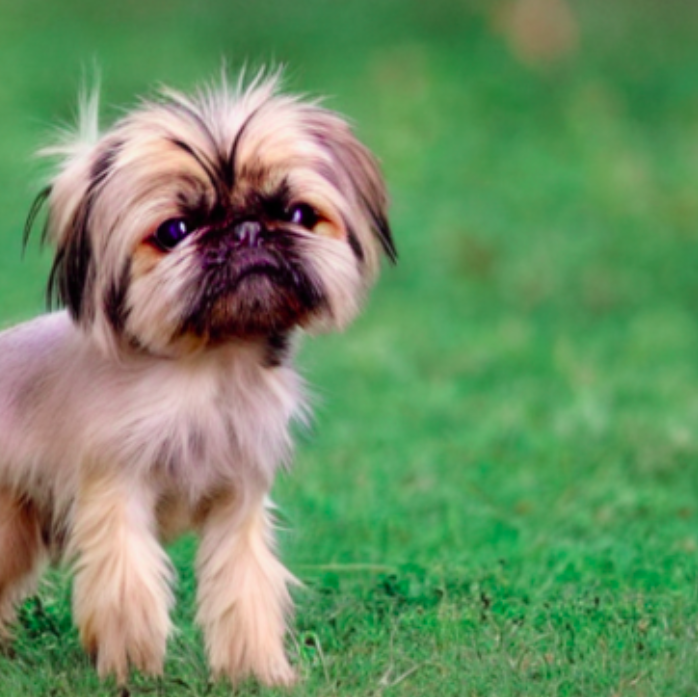} &
\includegraphics[width=\linewidth]{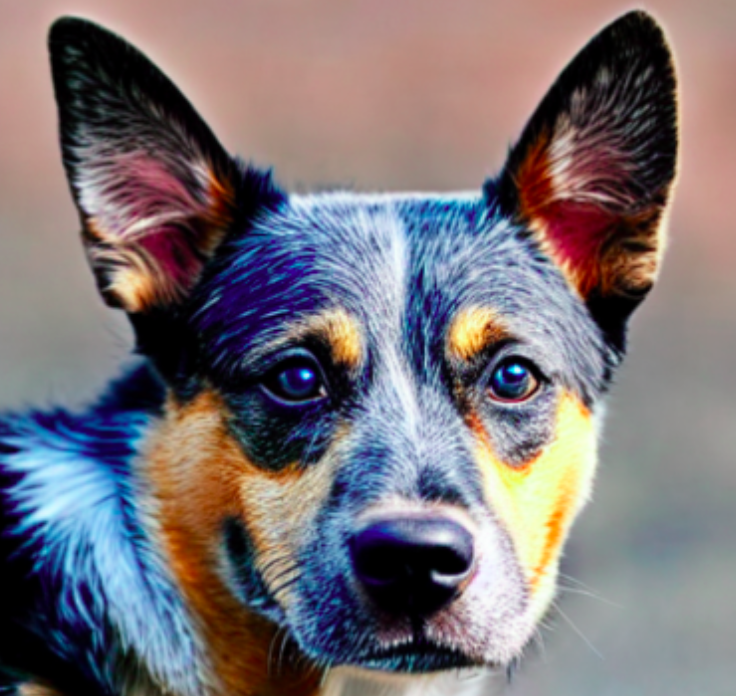} &
\includegraphics[width=\linewidth]{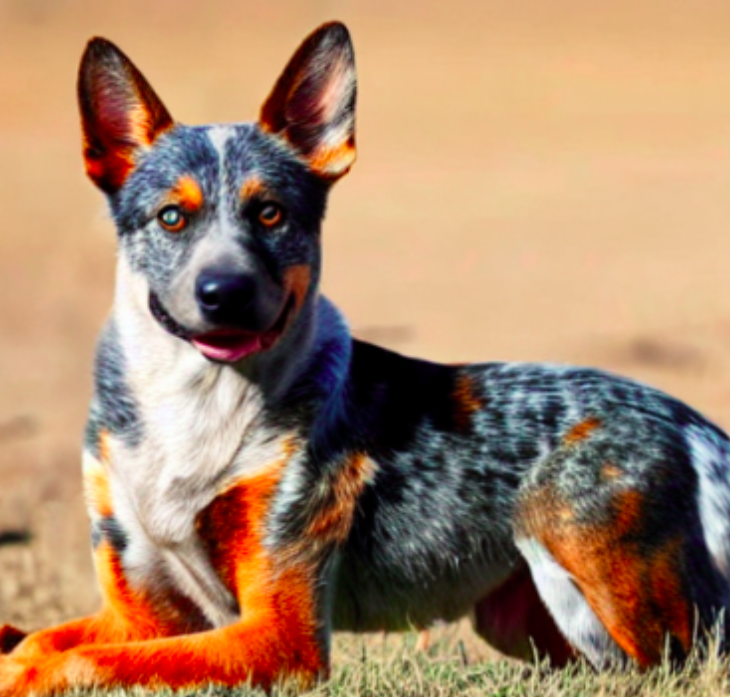} \\
Scene &
\makecell{Waterfall \\Cascade} &
Moon &
\includegraphics[width=\linewidth]{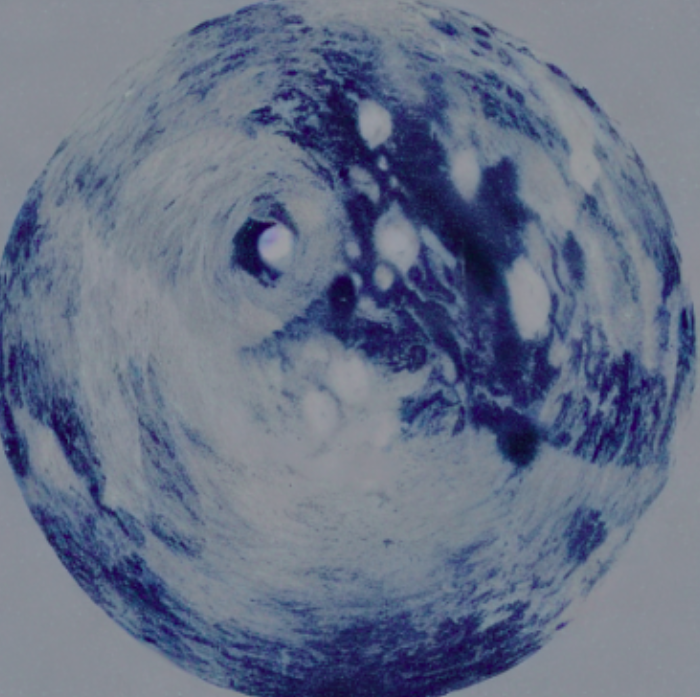} &
\includegraphics[width=\linewidth]{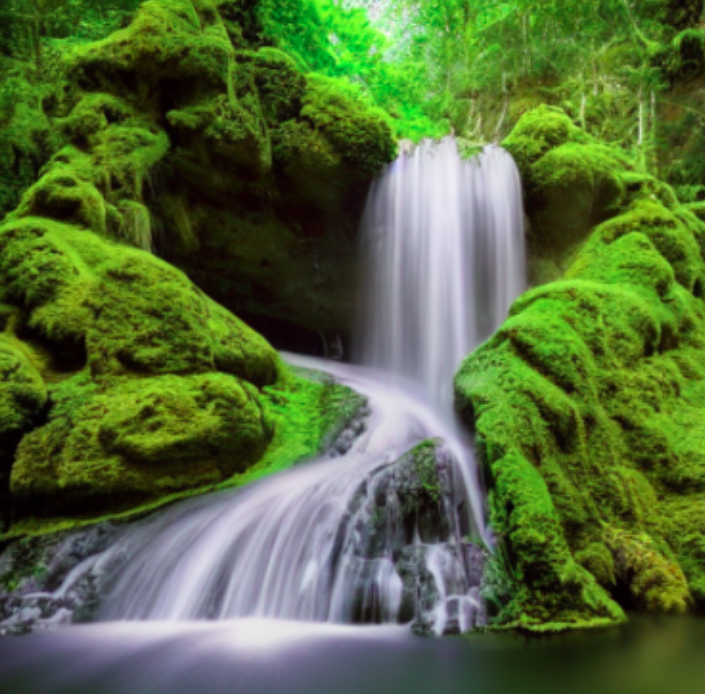} &
\includegraphics[width=\linewidth]{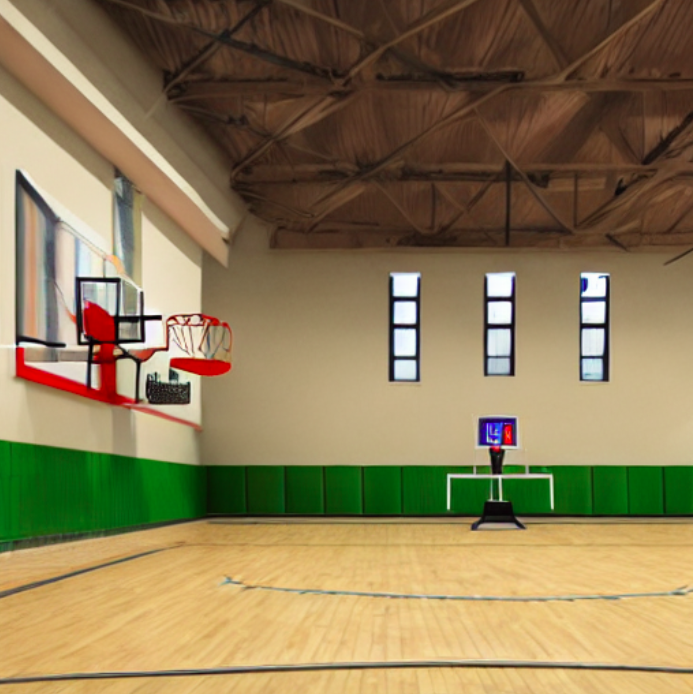} &
\includegraphics[width=\linewidth]{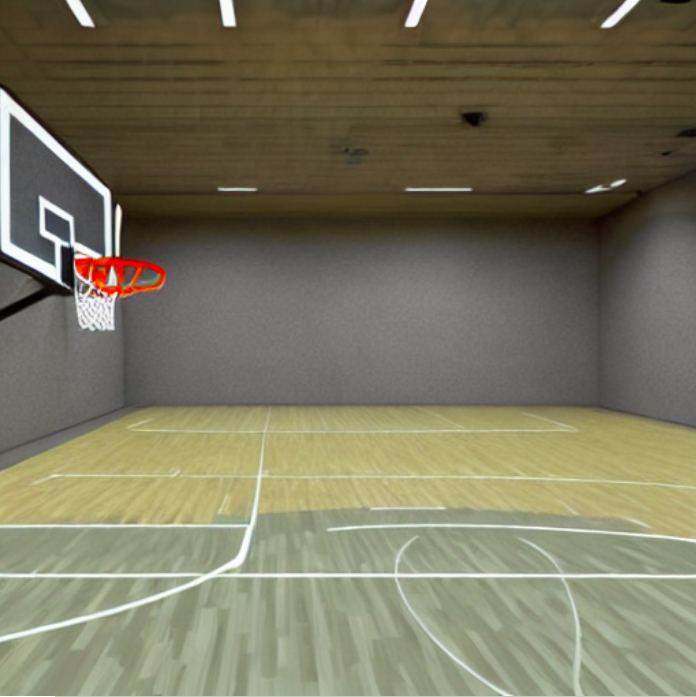} \\
\hline
\end{tabular}

\caption{Qualitative results for forgetting and retention in 3 selected tasks.}
\label{tab:qualitative}
\end{table}

\subsection{Qualitative results}\label{qualitative-results}
We performed unlearning in several tasks to showcase the applicability of SPARE in different contexts.
Across all tasks, and across several identities of each task and several concepts to be used for overwriting, SPARE demonstrated impressive unlearning. A few selected examples are shown in \cref{tab:qualitative}, unlearning concepts from the datasets Labeled Faces in the Wild (LFW) \cite{dataset_lfw}, AtharvaTaras Dog Breeds Dataset \cite{dataset_taras_dog_breeds}, and SUN Attributes \cite{dataset_sun_attributes}.

\section{Conclusion}
This paper presents SPARE, a method for unlearning concepts in text-to-image stable diffusion models. Leveraging on LoRA adaptation, the model provides lightweight modules that enable the removal of the forget concept. The evaluations in UnlearnCanvas benchmark showed that our method improves upon the current State-of-the-Art methods.

It is important to point out that our evaluations are restricted to a few datasets such as Imagenette and UnlearnCanvas. Due to our method's characteristic of overwriting a concept by another, the results might be sensitive to the choice of overwrite concept, a hypothesis which was not yet validated. Furthermore, the quantitative evaluations on UnlearnCanvas were restricted to unlearning performance, and further testing are needed in order to assess the method's robustness to sequential unlearning, knowledge revival, adversarial attacks, and comparable efficiency metrics.

The exploration on other benchmarks, different models, and additional tasks such as style editing or image classification could promote broader adoption of the method.
Explorations of the impact of overwriting choice as well as an automatic selection of concept would further enhance the capabilities of SPARE. Lastly, SPARE could theoretically be applied to architectures other than Stable Diffusion, but rigorous evaluation in that regard still needs to be performed.

\section*{Acknowledgements}
We are grateful to Fabien Baldacci and Marcos Escudero for reviewing our work throughout its development, and to Soham Vaze for his contributions to the Vision-Unlearning open-source library.

%
%
\bibliographystyle{splncs04}
\bibliography{main}

\clearpage
\appendix
\section{Hyperparameter configuration for UnlearnCanvas}\label{app:hyperparametersUC}

The hyperparameters used to unlearn the models during UnlearnCanvas evaluation were empirically explored during experiments to develop the method. They are related to the LoRA adapter trained for unlearning, scheduler type, optimizer, learning rate, among others, as described bellow.

\begin{itemize}
    \item Adam optimizer with beta1 = 0.9, beta 2 = 0.999, epsilon = 1e-08 and weigth decay = 1e-2.
    \item Learning Rate of 1e-4 with 0 warmup steps.
    \item Data augmentation with horizontal random flip.
    \item Batch size of 4.
    \item Maximum training steps of 1550.
    \item Constant scheduler.
    \item LoRA initial weights from Gaussian distribution, with dropout = 0.2, alpha = 8 and rank = 4.
    \item LoRA sparsity = 0\%
    \item Loss weighting: $\lambda_{forget} = 0.3$ and $\lambda_{retain} = 1.0$
\end{itemize}

\section{Impact of Sparsity}


\noindent To further evaluate the impact of gradient-based saliency masking on LoRA parameters, we conducted a series of experiments comparing full dense LoRA fine-tuning with the masked sparse LoRA configurations using our SPARE framework. We made the comparison on the following three distinct scenarios:

\begin{itemize}
    \item \textbf{Normal Style and Class Erasure on UnlearnCanvas:} Evaluating the removal of 10 specific styles and classes using the UnlearnCanvas benchmark.
    \item \textbf{Style Forgetting with out of Domain Retain Set:} Assessing style erasure performance when the retain set contains no other style images.
    \item \textbf{Unlearning with Almost no Retain:} Testing unlearning stability and performance in the near-absence of a retain set, which is done by weighting the retain loss by $0.02$.
\end{itemize}

\subsection{Results on Class and Style Forgetting}

In this experiment, we run unlearning with SPARE using $0\%$ sparsity and $50\%$ sparsity on the LoRAs for unlearning 10 styles and 10 classes from the UnlearnCanvas benchmark. The unlearning is run in both cases just for 500 steps with the same Hyperparameter used for the main UnlearnCanvas evaluation apart from the sparsity. The results as stated in table \ref{tab:unified_unlearning} shows that the sparse LoRA which updates only half the parameters than the normal dense LoRA performs better in UA, IRA and CRA for style unlearning, and on UA and CRA for class unlearning while also being faster.

\begin{table}[h!]
\centering
\caption{Comparison of Performance Metrics in UnlearnCanvas: SPARE with 50\% Sparse LoRA vs Dense LoRA.}
\label{tab:unified_unlearning}
\resizebox{\textwidth}{!}{
\begin{tabular}{l|ccc|ccc}
\hline
\multirow{2}{*}{\textbf{Unlearning Target}} & \multicolumn{3}{c|}{\textbf{50\% Sparse LoRA}} & \multicolumn{3}{c}{\textbf{Dense LoRA}} \\
\cline{2-7}
 & \textbf{UA} ($\uparrow$) & \textbf{IRA} ($\uparrow$) & \textbf{CRA} ($\uparrow$) & \textbf{UA} ($\uparrow$) & \textbf{IRA} ($\uparrow$) & \textbf{CRA} ($\uparrow$) \\
\hline
\multicolumn{7}{c}{\textbf{Style Unlearning}} \\
\hline
Artist Sketch & 100.00\% & 98.00\% & 93.40\% & 94.00\% & 98.22\% & 94.20\% \\
Bricks        & 100.00\% & 99.11\% & 96.20\% & 100.00\% & 96.89\% & 95.00\% \\
Byzantine     & 100.00\% & 96.67\% & 95.59\% & 94.00\% & 98.00\% & 95.80\% \\
Cartoon       & 100.00\% & 99.33\% & 93.79\% & 100.00\% & 98.89\% & 95.00\% \\
Cold Warm     & 100.00\% & 98.22\% & 95.80\% & 100.00\% & 98.67\% & 94.20\% \\
Color Fantasy & 100.00\% & 97.33\% & 96.20\% & 100.00\% & 97.56\% & 96.20\% \\
Comic Etch    & 100.00\% & 98.67\% & 92.80\% & 100.00\% & 98.00\% & 95.60\% \\
Crayon        & 100.00\% & 98.22\% & 97.60\% & 100.00\% & 97.11\% & 96.20\% \\
Cubism        & 100.00\% & 96.22\% & 96.20\% & 100.00\% & 98.22\% & 94.00\% \\
Dadaism       & 100.00\% & 98.67\% & 95.60\% & 100.00\% & 98.44\% & 94.20\% \\
\hline
\textbf{Style Average} & \textbf{100.00\%} & \textbf{98.04\%} & \textbf{95.32\%} & \textbf{98.40\%} & \textbf{98.00\%} & \textbf{95.04\%} \\
\hline
\multicolumn{7}{c}{\textbf{Object Unlearning}} \\
\hline
Architectures & 82.00\% & 97.78\% & 98.80\% & 100.00\% & 95.78\% & 98.40\% \\
Bears         & 98.00\% & 95.55\% & 99.40\% & 100.00\% & 95.11\% & 98.00\% \\
Birds         & 100.00\% & 94.44\% & 97.00\% & 96.00\% & 95.78\% & 98.60\% \\
Butterfly     & 86.00\% & 92.89\% & 97.80\% & 68.00\% & 94.89\% & 98.40\% \\
Cats          & 100.00\% & 93.33\% & 98.60\% & 100.00\% & 92.89\% & 99.00\% \\
Dogs          & 100.00\% & 93.78\% & 98.99\% & 100.00\% & 95.33\% & 98.80\% \\
Fishes        & 98.00\% & 93.33\% & 97.80\% & 92.00\% & 91.33\% & 98.60\% \\
Flame         & 100.00\% & 96.22\% & 97.80\% & 98.00\% & 94.89\% & 98.40\% \\
Flowers       & 90.00\% & 94.67\% & 99.40\% & 96.00\% & 94.00\% & 99.20\% \\
Frogs         & 98.00\% & 94.89\% & 98.80\% & 98.00\% & 95.56\% & 98.80\% \\
\hline
\textbf{Object Average} & \textbf{95.20\%} & \textbf{94.69\%} & \textbf{98.44\%} & \textbf{94.80\%} & \textbf{94.56\%} & \textbf{98.62\%} \\
\hline
\end{tabular}
}
\end{table}

\subsection{Results with Out-of-Domain Retain Data for Style Unlearning}

In this section, we compare SPARE with $0\%$ and $50\%$ sparsity when forgetting 10 styles from the UnlearnCanvas benchmark. Unlike previous experiments, we did not use other styles to compute the retain loss. Instead, we used 10 classes from the Imagenette dataset, which serves as an out-of-domain retain set for style unlearning. Training was conducted for 500 steps using the same hyperparameters used in \ref{app:hyperparametersUC}. 

The results in Table \ref{tab:style_unlearning_imagenet} show that both methods achieved $100\%$ UA. Regarding the retention of other styles, the sparse LoRA models demonstrate slightly better overall performance. However, if we analyze the retention for each unlearning session individually, we observe that sparse LoRAs perform much better in some cases, while dense LoRAs perform better in others. A possible reason is that when forgetting a specific style using gradient-based saliency to select relevant weights, we may also affect styles that are similar to the one being unlearned. This hurts the retention of those related styles, leading to worse performance compared to dense LoRAs.

One future direction to improve this gradient-based weight saliency is to factor the retain data into the weight selection process. If weight selection is based purely on importance to the forget data, it is unclear whether those same weights are also critical to the retain data. To avoid this collateral damage, applying gradient-based weight selection to the retain set could help filter out the weights identified by the saliency on the forget set. This would prevent the modification of weights that are important for retaining other styles.

\begin{table}[h!]
\centering
\caption{Performance comparison of Dense LoRA vs. Sparse LoRA for Style Unlearning using out-of-domain data (Imagenette) for retention.}
\label{tab:style_unlearning_imagenet}
\resizebox{\textwidth}{!}{
\begin{tabular}{l|ccc|ccc}
\hline
\multirow{2}{*}{\textbf{Target Style}} & \multicolumn{3}{c|}{\textbf{Dense LoRA}} & \multicolumn{3}{c}{\textbf{Sparse LoRA}} \\
\cline{2-7}
 & \textbf{UA} ($\uparrow$) & \textbf{Retain Style} ($\uparrow$) & \textbf{Retain Class} ($\uparrow$) & \textbf{UA} ($\uparrow$) & \textbf{Retain Style} ($\uparrow$) & \textbf{Retain Class} ($\uparrow$) \\
\hline
Artist Sketch & 100.00\% & 92.67\% & 96.80\% & 100.00\% & 93.78\% & 95.60\% \\
Bricks        & 100.00\% & 95.11\% & 96.60\% & 100.00\% & 94.44\% & 97.40\% \\
Byzantine     & 100.00\% & 96.89\% & 95.40\% & 100.00\% & 96.89\% & 95.60\% \\
Cartoon       & 100.00\% & 95.56\% & 95.60\% & 100.00\% & 96.89\% & 94.20\% \\
Cold Warm     & 100.00\% & 96.00\% & 93.00\% & 100.00\% & 95.78\% & 94.20\% \\
Color Fantasy & 100.00\% & 98.89\% & 96.60\% & 100.00\% & 98.89\% & 97.80\% \\
Comic Etch    & 100.00\% & 90.89\% & 95.80\% & 100.00\% & 88.00\% & 94.60\% \\
Crayon        & 100.00\% & 76.00\% & 98.40\% & 100.00\% & 86.00\% & 98.00\% \\
Cubism        & 100.00\% & 97.56\% & 94.60\% & 100.00\% & 92.89\% & 94.60\% \\
Dadaism       & 100.00\% & 80.89\% & 92.00\% & 100.00\% & 80.44\% & 92.20\% \\
\hline
\textbf{Average} & \textbf{100.00\%} & \textbf{92.00\%} & \textbf{95.50\%} & \textbf{100.00\%} & \textbf{92.40\%} & \textbf{95.40\%} \\
\hline
\end{tabular}
}
\end{table}



\subsection{Results for Style Unlearning with Minimal Retain Data}

Here, we repeated the comparison between SPARE with $0\%$ sparsity and $50\%$ sparsity when forgetting 10 styles from the UnlearnCanvas benchmark. However, we weighted the retain loss by a small factor ($\lambda_{\text{retain}} = 0.02$) to simulate a scenario where almost no retain data is available. As shown in Table \ref{tab:style_unlearning_small_weight}, the sparse LoRA once again performs slightly better than the dense LoRA across all three metrics: UA, IRA, and CRA.

\begin{table}[h!]
\centering
\caption{Performance comparison of Dense LoRA vs. Sparse LoRA for Style Unlearning (with $\lambda_{retain} = 0.02$).}
\label{tab:style_unlearning_small_weight}
\resizebox{\textwidth}{!}{
\begin{tabular}{l|ccc|ccc}
\hline
\multirow{2}{*}{\textbf{Style}} & \multicolumn{3}{c|}{\textbf{Dense LoRA}} & \multicolumn{3}{c}{\textbf{Sparse LoRA}} \\
\cline{2-7}
 & \textbf{UA} ($\uparrow$) & \textbf{Retain Style} ($\uparrow$) & \textbf{Retain Class} ($\uparrow$) & \textbf{UA} ($\uparrow$) & \textbf{Retain Style} ($\uparrow$) & \textbf{Retain Class} ($\uparrow$) \\
\hline
Artist Sketch & 100.00\% & 93.11\% & 97.00\% & 100.00\% & 92.89\% & 97.00\% \\
Bricks        & 100.00\% & 92.89\% & 97.60\% & 100.00\% & 94.22\% & 97.00\% \\
Byzantine     & 90.00\% & 94.89\% & 95.20\% & 96.00\% & 95.56\% & 96.00\% \\
Cartoon       & 100.00\% & 96.44\% & 94.60\% & 100.00\% & 96.67\% & 95.60\% \\
Cold Warm     & 100.00\% & 96.22\% & 94.20\% & 100.00\% & 95.33\% & 93.20\% \\
Color Fantasy & 100.00\% & 98.22\% & 98.20\% & 100.00\% & 97.33\% & 97.60\% \\
Comic Etch    & 100.00\% & 88.44\% & 96.20\% & 100.00\% & 88.00\% & 96.40\% \\
Crayon        & 100.00\% & 64.44\% & 98.60\% & 100.00\% & 66.44\% & 98.80\% \\
Cubism        & 100.00\% & 96.44\% & 94.60\% & 100.00\% & 96.22\% & 96.60\% \\
Dadaism       & 100.00\% & 77.11\% & 93.00\% & 100.00\% & 76.44\% & 92.00\% \\
\hline
\textbf{Average} & \textbf{99.00\%} & \textbf{89.80\%} & \textbf{95.90\%} & \textbf{99.60\%} & \textbf{89.90\%} & \textbf{96.00\%} \\
\hline
\end{tabular}
}
\end{table}

\section{Impact of the Replacing Concept}

Since SPARE depends on overwriting the concept to be unlearned with a surrogate safe concept, it is important to analyze how the choice of the replacing concept affects unlearning performance. To evaluate this, we experimented with SPARE on the Imagenette dataset by forgetting the ``gas pump'' class. We substituted it with the other nine classes from the dataset and compared both the quality of the unlearned model and the speed of unlearning in each scenario.

Our hypothesis was that replacing a concept with a safe concept that is ``close'' to the one being unlearned is better in terms of unlearning speed and overall retention quality. This approach should result in less destructive gradients, as the required weight changes would be smaller. As shown in Table \ref{tab:checkpoint_classifications}, when unlearning ``gas pump'' by overwriting it with other concepts for 500 steps of fine-tuning using SPARE, the speed at which the generated images transition to the new concept differs vastly depending on the replacing concept.

We calculated the FID between the set of images we have for ``gas pump'' and the images for each of the other nine classes. The ``truck'' class had the smallest FID with ``gas pump''. When we examine the percentage of generated images still classified as ``gas pump'' at various checkpointing steps, this percentage decreases the fastest when replacing the concept with ``truck''. Conversely, if we look at the results for ``chainsaw'', which has the highest FID with ``gas pump'', it exhibited the slowest decreasing trajectory. Furthermore, after 500 steps of fine-tuning, it only achieved $42\%$ unlearning. 

\begin{figure}[h!]
    \centering
    \includegraphics[width=0.8\textwidth]{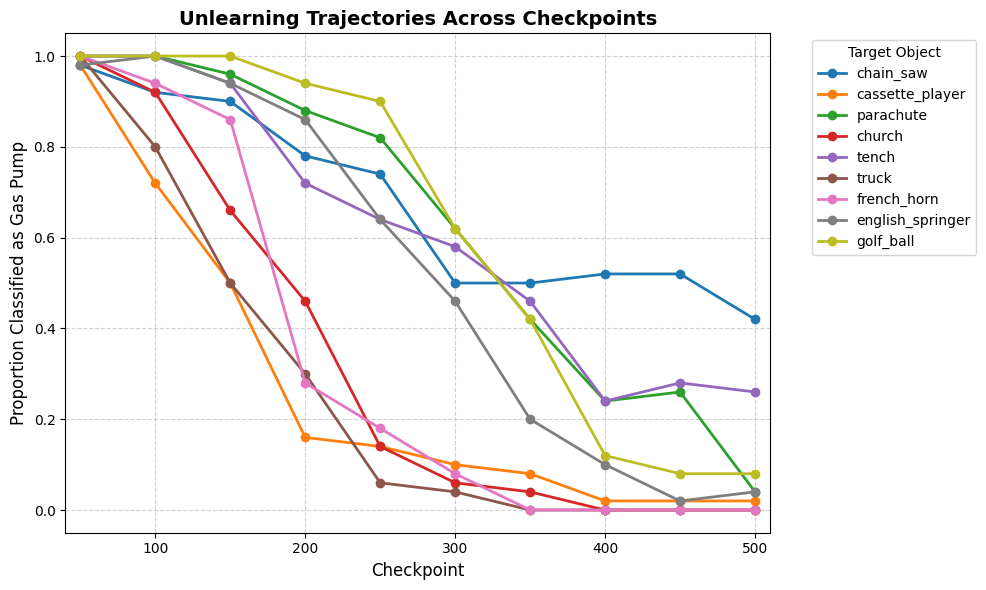}
    \caption{Unlearning trajectories when overwriting the target concept (``gas pump'') with nine different replacing concepts. The y-axis represents the percentage of generated images still classified as the target concept, illustrating the decay of the original concept over the course of the fine-tuning steps.}
    \label{fig:unlearning_trajectories}
\end{figure}

Figure \ref{fig:unlearning_trajectories} illustrates how the percentage of images classified as ``gas pump'' decreases for different overriding concepts at various checkpointing steps. We can observe that ``truck'' and ``cassette player'' exhibit the sharpest decreases, and they are the concepts with the two lowest FID scores with ``gas pump''. Additionally, ``chainsaw'' and ``tench'', which have the two highest FID scores with ``gas pump'', display the slowest trajectories.

\begin{table}[h!]
\centering
\scriptsize
\setlength{\tabcolsep}{5pt} 
\caption{Percentage of generated samples still classified as the target concept (``gas pump'') across checkpointing steps, alongside the FID between the target and replacing concepts. This illustrates how the visual similarity (measured by FID) of the replacing concept influences the unlearning dynamics over time.}
\label{tab:checkpoint_classifications}
\begin{tabular}{l|c|cccccccccc}
\hline
\multirow{2}{*}{\textbf{Replacing concept}} & \multirow{2}{*}{\textbf{FID}} & \multicolumn{10}{c}{\textbf{Checkpoints}} \\
\cline{3-12}
 &  & \textbf{50} & \textbf{100} & \textbf{150} & \textbf{200} & \textbf{250} & \textbf{300} & \textbf{350} & \textbf{400} & \textbf{450} & \textbf{500} \\
\hline
chain\_saw        & 346.58 & 98  & 92  & 90  & 78 & 74 & 50 & 50 & 52 & 52 & 42 \\
cassette\_player  & 200.47 & 98  & 72  & 50  & 16 & 14 & 10 & 8  & 2  & 2  & 2  \\
parachute         & 211.51 & 100 & 100 & 96  & 88 & 82 & 62 & 42 & 24 & 26 & 4  \\
church            & 211.27 & 100 & 92  & 66  & 46 & 14 & 6  & 4  & 0  & 0  & 0  \\
tench             & 228.68 & 100 & 100 & 94  & 72 & 64 & 58 & 46 & 24 & 28 & 26 \\
truck             & \textbf{191.98} & 100 & 80  & 50  & 30 & 6  & 4  & 0  & 0  & 0  & 0  \\
french\_horn      & 218.29 & 100 & 94  & 86  & 28 & 18 & 8  & 0  & 0  & 0  & 0  \\
english\_springer & 227.02 & 98  & 100 & 94  & 86 & 64 & 46 & 20 & 10 & 2  & 4  \\
golf\_ball        & 220.27 & 100 & 100 & 100 & 94 & 90 & 62 & 42 & 12 & 8  & 8  \\
\hline
\end{tabular}
\end{table}

\section{Further Analysis on Timestep Targeting}

One argument we presented for why our timestep targeting approach makes unlearning faster is that it focuses the process on the most critical timesteps. These are the timesteps where the two concepts diverge according to the model's internal understanding. In the main text, we presented empirical evidence demonstrating that our proposed sampling strategy accelerates unlearning. Here, we provide qualitative evidence explaining why this occurs, using an unlearning session where the model forgets ``Colin Powell'' by overwriting him with ``Barack Obama''.

The suggested timesteps to target for this unlearning session were from 100 to 650. Because these are later timesteps, it indicates that to replace Colin Powell with Barack Obama, the model only needs to focus on refining details. It does not need to alter earlier timesteps where global features like shape and general color are established. Therefore, when comparing images generated by the unlearned model to those from the original model, we expect the visual changes to be minimal and detail-oriented when using the late-timestep sampling scheme.

\begin{figure}[!ht]
    \centering
    \begin{subfigure}[b]{\textwidth}
        \centering
        \includegraphics[width=\textwidth]{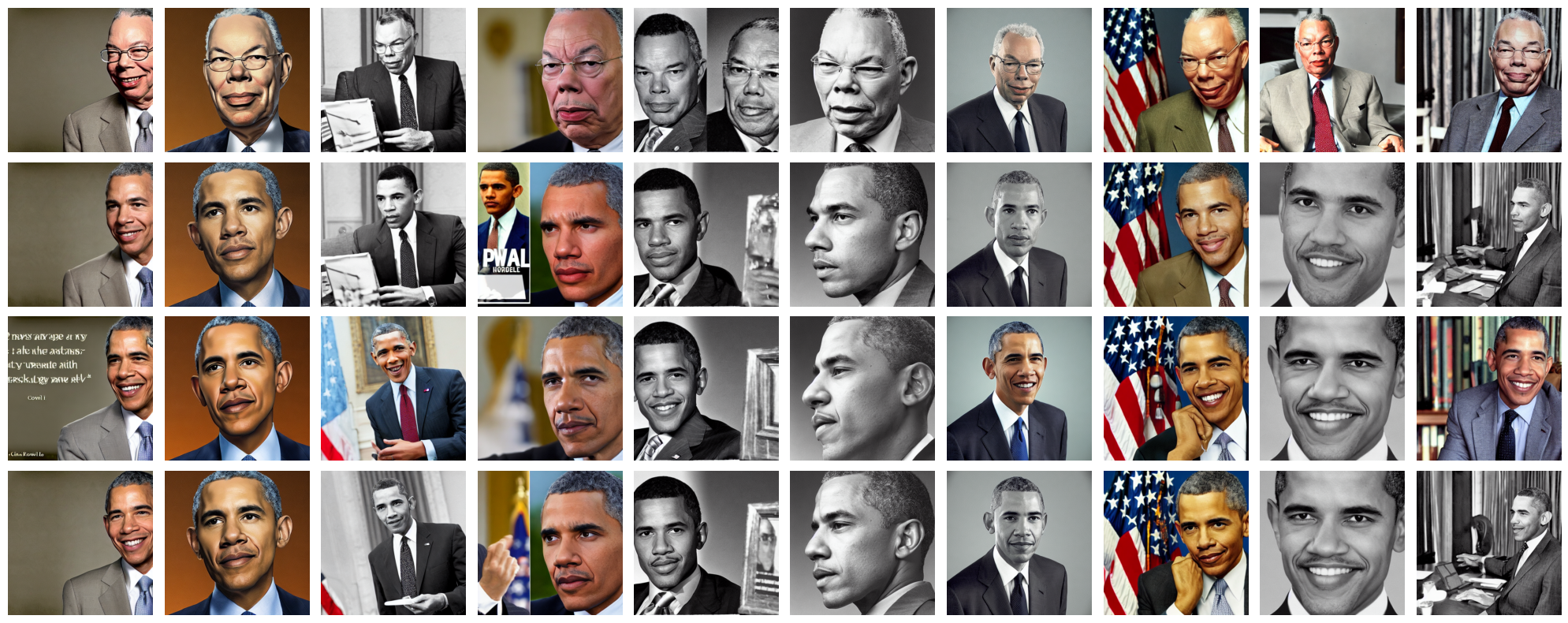}
        \caption{Images generated for the prompt ``Colin Powell''. The rows, from top to bottom, correspond to the original model, followed by the 100th, 200th, and 300th checkpoints during fine-tuning using \textbf{uniform} timestep sampling.}
        \label{fig:top_img2}
    \end{subfigure}
    
    \vspace{0.2cm} 
    
    \begin{subfigure}[b]{\textwidth}
        \centering
        \includegraphics[width=\textwidth]{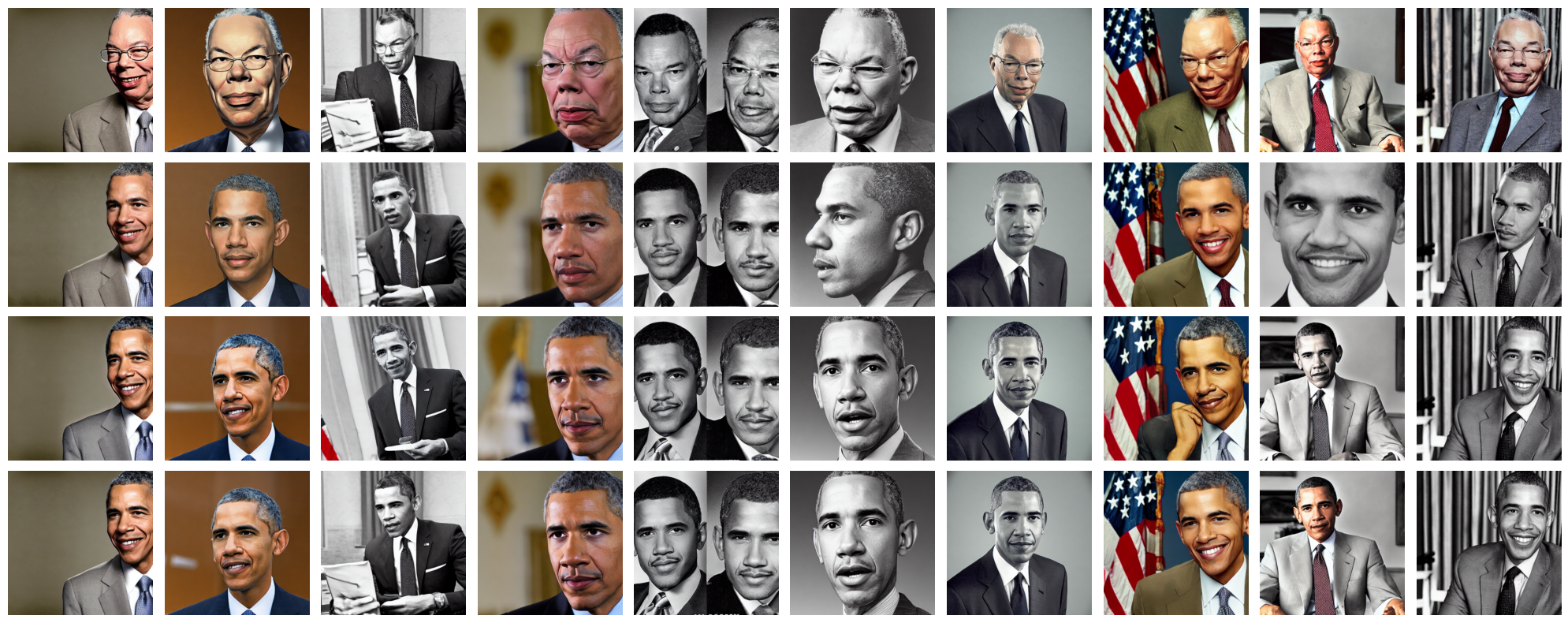}
        \caption{Images generated for the prompt ``Colin Powell''. The rows, from top to bottom, correspond to the original model, followed by the 100th, 200th, and 300th checkpoints during fine-tuning using our proposed \textbf{late} timestep sampling.}
        \label{fig:bottom_img2}
    \end{subfigure}
    
    \caption{\textbf{Comparison of visual changes during unlearning with uniform versus late timestep sampling.} The figure displays images generated for the prompt ``Colin Powell'' during an unlearning session that overwrites him with ``Barack Obama''. \ref{fig:top_img2} shows that uniform timestep sampling introduces unnecessary alterations to global features, such as pose, face-to-image proportion, and overall color. In contrast, \ref{fig:bottom_img2} shows the proposed late timestep sampling focuses on refining specific facial details, resulting in minimal changes to the original pose and color structure. This confirms that late sampling accelerates unlearning by avoiding unnecessary modifications.}
    \label{fig:powel_obamaaaa}
\end{figure}

Figure \ref{fig:powel_obamaaaa} clearly illustrates this effect. Although the replacement of Colin Powell with Barack Obama is successful in both scenarios, using uniform sampling (Figure \ref{fig:top_img2}) introduces unnecessary changes in pose, face-to-image proportion, and color. Conversely, Figure \ref{fig:bottom_img2} shows that using our proposed late sampling scheme results in minimal changes to pose and overall color. This demonstrates that our targeted timestep sampling scheme makes unlearning faster by not wasting fine-tuning steps on irrelevant timesteps.

    
    
    

\section{Validating Concept Substitution via Cross-
Attention}


To verify the internal mechanism of our unlearning algorithm, we move beyond evaluating fully generated images and analyze the model's intermediate representations—specifically, the cross-attention layers in the diffusion model's U-Net. These layers align spatial image features with specific tokens from the text prompt. Visualizing these maps reveals exactly which image regions the model associates with a concept, providing transparency into how the model interprets the token being unlearned.

To generate attention maps between text and image tokens, we extracted cross-attention values from the $16 \times 16$ resolution blocks of the U-Net. Because cross-attention at this resolution occurs five times per forward pass (twice during downsampling, thrice during upsampling), we averaged the maps across these five instances and across all attention heads. We ignored attention maps tied to the unconditional prompt used for classifier-free guidance. Finally, we averaged the maps across all diffusion timesteps to produce a single final attention map for each generated image.


The efficacy of our safe concept replacement strategy is evident when comparing attention maps before and after unlearning. We illustrate this using a model fine-tuned to forget ``gas pump'' by overwriting it with ``garbage truck''. As shown in Figure \ref{fig:attention_comparison}, the ``gas pump'' token embedding in the original model strongly attends to regions representing a gas pump. After fine-tuning, the same token instead attends to pixels belonging to a garbage truck. This difference is visible even at the first denoising step—before any semantic information emerges from the image tokens—indicating that the modified cross-attention is the primary driver of the model's behavioral change. This confirms our algorithm successfully rewrites the semantic mapping in the model's weights, rather than merely suppressing the original concept. Figure \ref{fig:across} further details how the attention map for the ``gas pump'' token evolves over the fine-tuning steps.


\addtocounter{figure}{1} 

\begin{figure}[b] 
    \centering
    \includegraphics[width=0.99\linewidth]{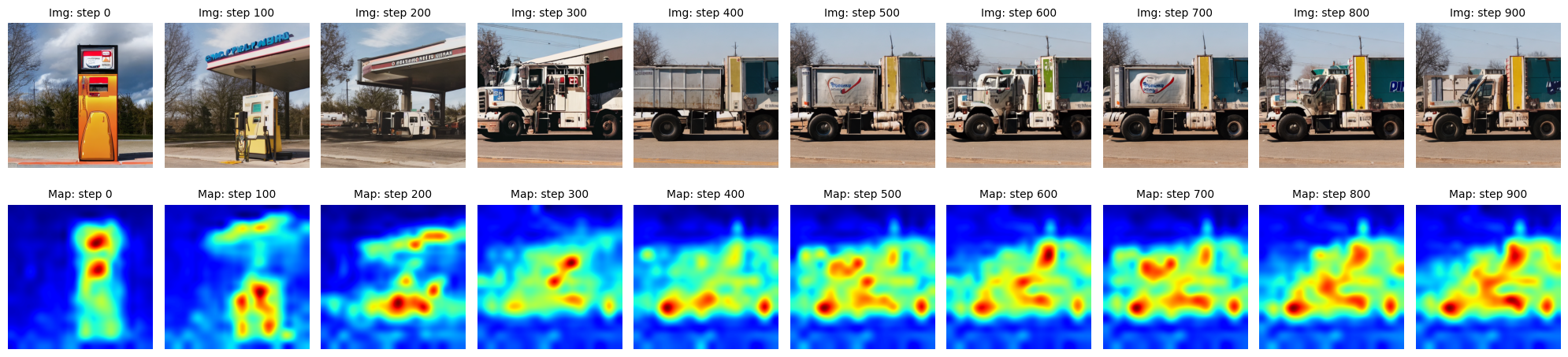}
    \caption{Attention map evolution across fine-tuning steps. The figure illustrates the attention map of the ``gas pump'' token over the generated image tokens. As training progresses, the token increasingly attends to pixels representing the replacing concept (garbage truck).}
    \label{fig:across}
\end{figure}

\addtocounter{figure}{-2}

\begin{figure}[htpb]
    \centering
    \begin{subfigure}[t]{0.95\textwidth}
        \centering
        \includegraphics[width=\linewidth]{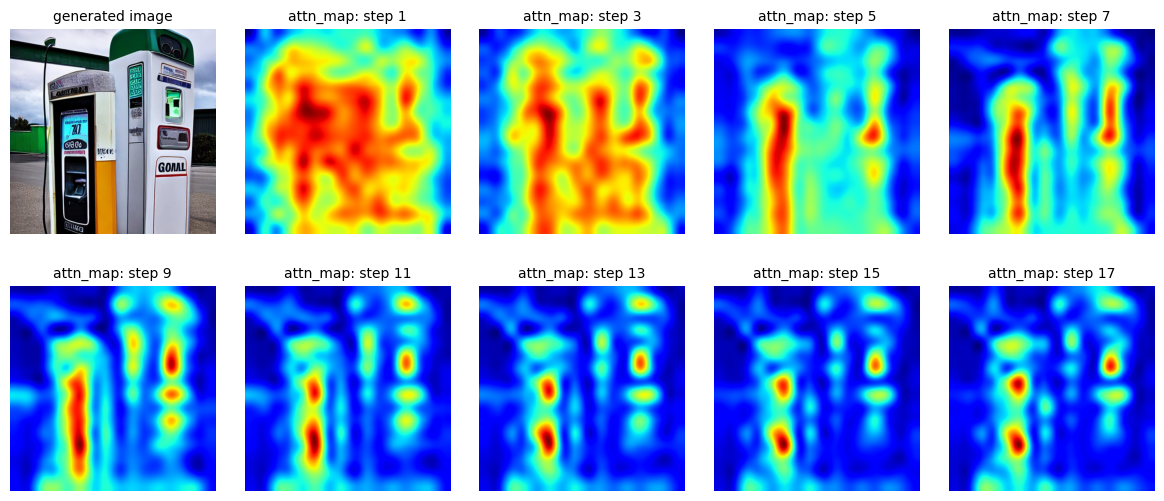}
        \caption{Attention maps over diffusion timesteps (original model)}
        \label{fig:attn_orig_timesteps}
    \end{subfigure}

    \vspace{0.8em}

    \begin{subfigure}[t]{0.95\textwidth}
        \centering
        \includegraphics[width=\linewidth]{figures/fine_timesteps.png}
        \caption{Attention maps over diffusion timesteps (fine-tuned model)}
        \label{fig:attn_ft_timesteps2}
    \end{subfigure}

    \vspace{1em}

    \begin{subfigure}[t]{0.48\textwidth}
        \centering
        \includegraphics[width=\linewidth]{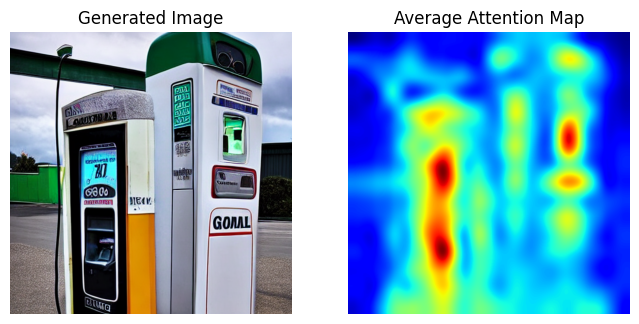}
        \caption{Mean attention map (original model)}
        \label{fig:attn_orig_mean}
    \end{subfigure}
    \hfill
    \begin{subfigure}[t]{0.48\textwidth}
        \centering
        \includegraphics[width=\linewidth]{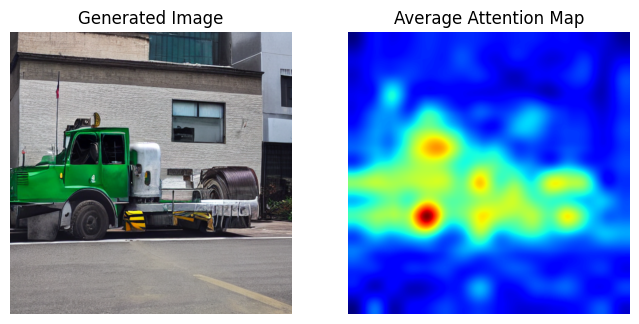}
        \caption{Mean attention map (fine-tuned model)}
        \label{fig:attn_ft_mean}
    \end{subfigure}

    \caption{\textbf{Comparison of cross-attention maps for the ``gas pump'' token before and after unlearning.} The figure illustrates how the attention between the ``gas pump'' token embedding and the generated image tokens changes when the original Stable Diffusion model is fine-tuned to overwrite the concept with a truck. Subfigures (a) and (b) display the evolution of attention maps across individual diffusion timesteps for the original and fine-tuned models, respectively. Subfigures (c) and (d) present the mean attention maps aggregated over all timesteps. The clear shift in the attended regions confirms that the internal semantic mapping of the target concept has been successfully rewritten.}
    \label{fig:attention_comparison}
\end{figure}


\end{document}

%% file: quantitative_eval.tex
To address the challenges of assessing \ac{MU} in image generation, we utilize the UnlearnCanvas benchmark \cite{Zhang2024UnlearnCanvas}. This high-resolution stylized dataset features dual supervision, providing annotated images with both style and object labels, to precisely isolate unlearning targets. This allows for the evaluation of ``in-domain'' retainability (preserving concepts within the same category) and ``cross-domain'' retainability (maintaining fidelity across different categories).

Our evaluation focused on four core metrics, from the unlearning performance set of UnlearnCanvas, using the pretrained ViT-Large classifiers and the Stable Diffusion (SD) v1.5 backbone:
\begin{itemize}
    \item \textbf{Unlearn Accuracy (UA):} Measured as 100\% minus the classifier accuracy of the forgotten concept.
    \item \textbf{In-domain Retain Accuracy (IRA):} Mean classifier accuracy for retained concepts within the same domain.
    \item \textbf{Cross-domain Retain Accuracy (CRA):} Mean accuracy for concepts in the alternative domain (e.g., object fidelity during style removal or vice versa).
    \item \textbf{FID:} The Fréchet Inception Distance between the generated images and the original dataset (excluding the unlearned concept).
\end{itemize}



The SPARE method was used with 0\% of sparsity in the UnlearnCanvas evaluation, the unlearning process considered a forget set of all images of the forget concept and randomly sampled images from other concepts to compose the retain set. The hyperparameters used to unlearn the models were empirically explored during experiments and are described in the supplemental materials \cref{app:hyperparametersUC}.

For each unlearned model, prompts were created for all possible style-object combinations, generating 5 images per combination with different seeds. While UnlearnCanvas reports runtime and peak memory on NVIDIA RTX A6000 GPUs, our metrics were recorded on NVIDIA V100 (16 GB) and A100 (40 GB) GPUs; thus, efficiency results are not directly comparable to the benchmark’s original reports.

A distinctive feature of SPARE is the use of an overwriting concept, which we defined as a random concept from the retain set in the same domain as the forget concept, for instance using "Meta Physics" to overwrite "Van Gogh". One unlearned model was generated for each concept in both style and object domains. Each unlearned model was generated using LoRA adaptation, resulting in a consistent memory overhead of 3.1MB per model when the adapter is not merged (case in which there would be no memory overhead).

\begin{figure}[h!]
    \centering
    \includegraphics[width=0.8\linewidth]{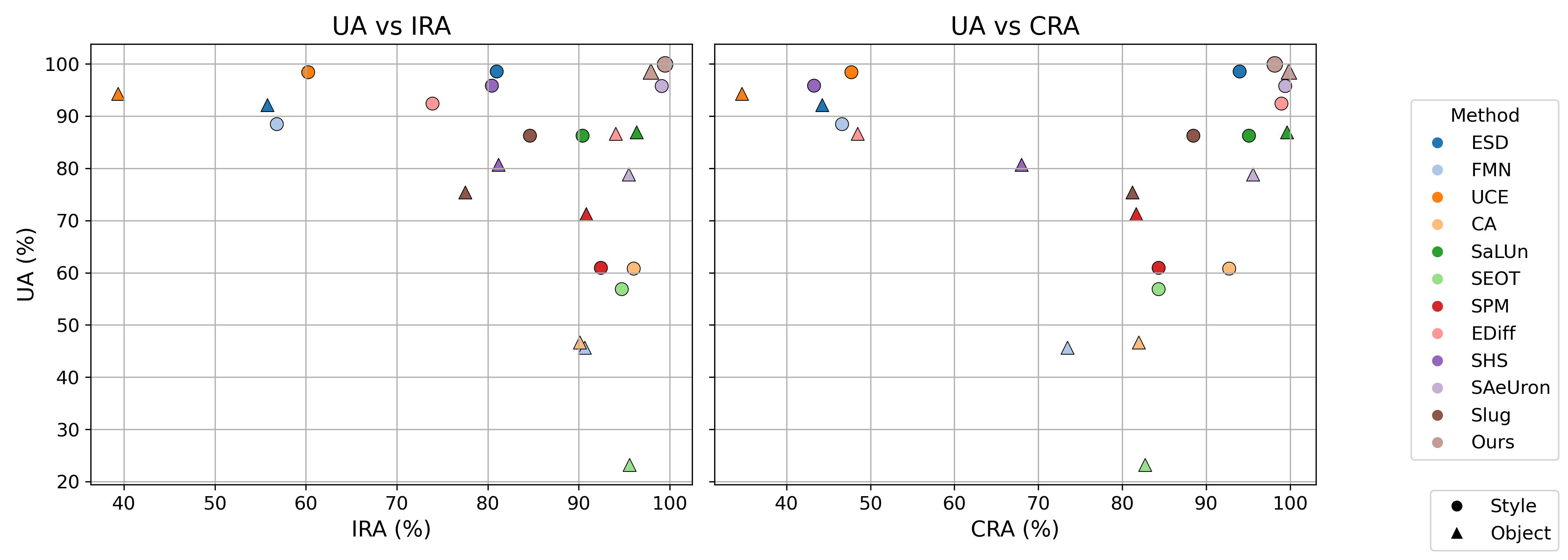}
    \caption{Unlearning accuracies by retaining accuracies for the methods shown in \cref{tab:performance}.}
    \label{fig:ua_vs_ira_cra}
\end{figure}

As shown in \cref{tab:performance}, SPARE achieves performance comparable to or exceeding current State-of-the-Art methods. This is further visualized in \cref{fig:ua_vs_ira_cra}, where SPARE consistently occupies the ideal region (high UA vs. high IRA/CRA). Notably, our method demonstrates a balanced response across both style and object domains.

\begin{table}[h!]
\centering
\caption{Performance Metrics in UnlearnCanvas. For each metric, the best scoring method is highlighted in \colorbox{mygreen}{green}, the second best in \colorbox{myyellow}{yellow}, and the worst in \colorbox{myred}{red}.}
\label{tab:performance}
\resizebox{\textwidth}{!}{%
\begin{tabular}{l|ccc|ccc|c|ccc}
\hline
\multirow{3}{*}{\textbf{Method}} & \multicolumn{7}{c|}{\textbf{Effectiveness}} & \multicolumn{3}{c}{\textbf{Efficiency}} \\
\cline{2-11}
 & \multicolumn{3}{c|}{\textbf{Style Unlearning}} & \multicolumn{3}{c|}{\textbf{Object Unlearning}} & \textbf{FID} & \textbf{Time} & \textbf{Memory} & \textbf{Storage} \\
\cline{2-7}
 & \textbf{UA} ($\uparrow$) & \textbf{IRA} ($\uparrow$) & \textbf{CRA} ($\uparrow$) & \textbf{UA} ($\uparrow$) & \textbf{IRA} ($\uparrow$) & \textbf{CRA} ($\uparrow$) & ($\downarrow$) & (s) ($\downarrow$) & (GB) ($\downarrow$) & (GB) ($\downarrow$) \\
\hline
ESD \cite{gandikota2023} & \cellcolor{myyellow} 98.58\% & 80.97\% & 93.96\% & 92.15\% & 55.78\% & 44.23\% & 65.55 & 6163 & 17.8 & \cellcolor{myred} 4.3 \\
FMN \cite{Zhang2023_fmn} & 88.48\% & \cellcolor{myred}56.77\% & 46.60\% & 45.64\% & 90.63\% & 73.46\% & 131.37 & 350 & 17.9 & 4.2 \\
UCE \cite{Gandikota2023UCE} & 98.40\% & 60.22\% & 47.71\% & \cellcolor{myyellow} 94.31\% & \cellcolor{myred}39.35\% & \cellcolor{myred}34.67\% & \cellcolor{myred} 182.01 & 434 & \cellcolor{mygreen}5.1 & 1.7 \\
CA \cite{kumari2023} & 60.82\% & 96.01\% & 92.70\% & 46.67\% & 90.11\% & 81.97\% & \cellcolor{mygreen}54.21 & 734 & 10.1 & 4.2 \\
SaLUn \cite{fan2023salun} & 86.26\% & 90.39\% & 95.08\% & 86.91\% & 96.35\% & \cellcolor{myyellow} 99.59\% & 61.05 & 667 & 30.8 & 4.0 \\
SEOT \cite{li2024SEOT} & \cellcolor{myred}56.90\% & 94.68\% & 84.31\% & \cellcolor{myred}23.25\% & \cellcolor{myyellow} 95.57\% & 82.71\% & 62.38 & \cellcolor{mygreen}95 & 7.34 & \cellcolor{mygreen}0.0 \\
SPM \cite{Lyu2023SPM} & 60.94\% & 92.39\% & 84.33\% & 71.25\% & 90.79\% & 81.65\% & 59.79 & \cellcolor{myred}29700 & 6.9 & 4.0 \\
EDiff \cite{Wu2024EraseDiff} & 92.42\% & 73.91\% & \cellcolor{myyellow}98.93\% & 86.67\% & 94.03\% & 48.48\% & 81.42 & 1567 & 27.8 & 4.0 \\
SHS \cite{Wu2024_scissorhands} & 95.84\% & 80.42\% & \cellcolor{myred}43.27\% & 80.73\% & 81.15\% & 67.99\% & 119.34 & 1223 & \cellcolor{myred}31.2 & 4.0 \\

SAeUron \cite{Cywinski2025}  & 95.80\% & \cellcolor{myyellow} 99.10\% & 99.40\% & 78.82\% & 95.47\% & 95.58\% & 94.03 & 62.15 & 2.8 & 0.2 \\

Slug \cite{Cai2025Slug}  & 86.29\% & 84.59\% & 88.43\% & 75.43\% & 77.50\% & 81.18\% & 75.97 & 39 & 3.61 & 0.04 \\

Ours & \cellcolor{mygreen} 99.96\% & \cellcolor{mygreen} 99.45\% & \cellcolor{myyellow} 98.16\% & \cellcolor{mygreen} 98.55\% & \cellcolor{mygreen} 97.88\% & \cellcolor{mygreen} 99.82\% & \cellcolor{myyellow} 54.36 & 2025.68* & 20.57* & \cellcolor{mygreen} 0.03 \\
\hline
\end{tabular}%
}
\end{table}


Overall, SPARE proves highly effective, though further evaluation is required to fully standardize efficiency metrics.

\subsection{Experiments with sparsity}

The comprehensive evaluation on UnlearnCanvas initially did not account for sparsity or feature localization when modifying the models via LoRA. To investigate the influence of parameter sparsity on unlearning performance, we conducted an ablation study on a targeted subset of objects and styles using a 50\% sparsity constraint. Notably, this experiment was performed under a reduced computational budget, using 500 training steps compared to the 1500 steps used in the full evaluation.
The results, in \cref{tab:sparse}, illustrate the performance across the selected concepts. Although the metrics show a marginal decline compared to the non-sparse models, the UA, IRA, and CRA remain superior to the values reported by the majority of competing state-of-the-art methods.

\begin{table}[h!]
\centering
\caption{Performance Metrics in subset of UnlearnCanvas with SPARE using 50\% sparsity.}
\label{tab:sparse}
\resizebox{\textwidth}{!}{
\begin{tabular}{c|ccc|c|ccc}
\hline
\multicolumn{8}{c}{\textbf{Effectiveness}} \\
\cline{1-8}
\multirow{2}{*}{\textbf{Style}} & \multicolumn{3}{c|}{\textbf{Style Unlearning}} & \multirow{2}{*}{\textbf{Object}} & \multicolumn{3}{c}{\textbf{Object Unlearning}} \\
\cline{2-4} \cline{6-8}
 & \textbf{UA} ($\uparrow$) & \textbf{IRA} ($\uparrow$) & \textbf{CRA} ($\uparrow$) & & \textbf{UA} ($\uparrow$) & \textbf{IRA} ($\uparrow$) & \textbf{CRA} ($\uparrow$) \\
\hline
Artist Sketch & 100.00\% & 98.00\% & 93.40\% & Architectures &  82.00\% & 97.78\% & 98.80\% \\ \hline
    Bricks    & 100.00\% & 99.11\% & 96.20\% &     Bears     &  98.00\% & 95.55\% & 99.40\% \\ \hline
  Byzantine   & 100.00\% & 96.67\% & 95.59\% &     Birds     & 100.00\% & 94.44\% & 97.00\% \\ \hline
   Cartoon    & 100.00\% & 99.33\% & 93.79\% &   Butterfly   &  86.00\% & 92.89\% & 97.80\% \\ \hline
  Cold Warm   & 100.00\% & 98.22\% & 95.80\% &      Cats     & 100.00\% & 93.33\% & 98.60\% \\ \hline
Color Fantasy & 100.00\% & 97.33\% & 96.20\% &      Dogs     & 100.00\% & 93.78\% & 98.99\% \\ \hline
  Comic Etch  & 100.00\% & 98.67\% & 92.80\% &     Fishes    &  98.00\% & 93.33\% & 97.80\% \\ \hline
    Crayon    & 100.00\% & 98.22\% & 97.60\% &     Flame     & 100.00\% & 96.22\% & 97.80\% \\ \hline
    Cubism    & 100.00\% & 96.22\% & 96.20\% &    Flowers    &  90.00\% & 94.67\% & 99.40\% \\ \hline
   Dadaism    & 100.00\% & 98.67\% & 95.60\% &     Frogs     &  98.00\% & 94.89\% & 98.80\% \\ \hline
    \textbf{Average}       & \textbf{100.00\%} & \textbf{98.04\%} & \textbf{95.32\%} & \textbf{Average}   &  \textbf{95.20\%} & \textbf{94.69\%} & \textbf{98.44\%}\\ \hline
\end{tabular}
}
\end{table}


While a full-scale benchmark is necessary for definitive confirmation, the sparse approach shows potential to achieve performance parity with its non-sparse counterpart while maintaining a significantly smaller parameter footprint. Furthermore, in scenarios where a large, in-domain retain dataset is available, the distillation objective alone is powerful enough to prevent catastrophic forgetting. Conversely, if a retain dataset is unavailable or comes from a completely different domain, identifying the optimal weights via sparse LoRA becomes critical. We will include detailed experiments addressing these scenarios in the supplementary material.

%% file: main.bib
@String(CVPR  = {IEEE Conf. Comput. Vis. Pattern Recog.})

@String(ICCV  = {Int. Conf. Comput. Vis.})

@String(ECCV  = {Eur. Conf. Comput. Vis.})

@String(NeurIPS = {Adv. Neural Inform. Process. Syst.})

@String(ICML  = {Int. Conf. Mach. Learn.})

@String(ICLR  = {Int. Conf. Learn. Represent.})

@String(CVPRW = {IEEE Conf. Comput. Vis. Pattern Recog. Worksh.})

@String(CVPR  = {CVPR})

@String(ICCV  = {ICCV})

@String(ECCV  = {ECCV})

@String(NeurIPS = {NeurIPS})

@String(ICML  = {ICML})

@String(ICLR  = {ICLR})

@String(CVPRW = {CVPRW})

@article{xu2023review,
author = {Xu, Heng and Zhu, Tianqing and Zhang, Lefeng and Zhou, Wanlei and Yu, Philip S.},
title = {Machine Unlearning: A Survey},
year = {2023},
issue_date = {January 2024},
publisher = {Association for Computing Machinery},
address = {New York, NY, USA},
volume = {56},
number = {1},
issn = {0360-0300},
url = {https://doi.org/10.1145/3603620},
doi = {10.1145/3603620},
journal = {ACM Comput. Surv.},
month = aug,
articleno = {9},
numpages = {36}
}

@misc{gdpr2016,
  author       = {{European Union}},
  title        = {{Regulation (EU) 2016/679 of the European Parliament and of the Council of 27 April 2016 on the protection of natural persons with regard to the processing of personal data and on the free movement of such data, and repealing Directive 95/46/EC (General Data Protection Regulation)}},
  howpublished = {\url{https://eur-lex.europa.eu/eli/reg/2016/679/oj}},
  note         = {OJ L 119, 4.5.2016, p. 1--88},
  year         = {2016}
}

@misc{imagenette,
  author       = {Howard, Jeremy},
  title        = {Imagenette},
  year         = {2019},
  howpublished = {\url{https://github.com/fastai/imagenette}},
  note         = {Accessed: 2025-05-21}
}

@article{song2020ddim,
  title={Denoising diffusion implicit models},
  author={Song, Jiaming and Meng, Chenlin and Ermon, Stefano},
  journal={arXiv preprint arXiv:2010.02502},
  year={2020}
}

@inproceedings{wu2018_memory_replay_gans,
 author = {Wu, Chenshen and Herranz, Luis and Liu, Xialei and wang, yaxing and van de Weijer, Joost and Raducanu, Bogdan},
 booktitle =NeurIPS,
 editor = {S. Bengio and H. Wallach and H. Larochelle and K. Grauman and N. Cesa-Bianchi and R. Garnett},
 pages = {},
 publisher = {Curran Associates, Inc.},
 title = {Memory Replay GANs: Learning to Generate New Categories without Forgetting},
 url = {https://proceedings.neurips.cc/paper_files/paper/2018/file/a57e8915461b83adefb011530b711704-Paper.pdf},
 volume = {31},
 year = {2018}
}

@INPROCEEDINGS{dataset_sun_attributes,
  author={Patterson, Genevieve and Hays, James},
  booktitle=CVPR,
  title={SUN attribute database: Discovering, annotating, and recognizing scene attributes}, 
  year={2012},
  volume={},
  number={},
  pages={2751-2758},
  doi={10.1109/CVPR.2012.6247998},
  ISSN={1063-6919},
  month={June},
}

@inproceedings{dataset_lfw,
  TITLE = {{Labeled Faces in the Wild: A Database forStudying Face Recognition in Unconstrained Environments}},
  AUTHOR = {Huang, Gary B. and Mattar, Marwan and Berg, Tamara and Learned-Miller, Eric},
  URL = {https://inria.hal.science/inria-00321923},
  BOOKTITLE = {{Workshop on Faces in 'Real-Life' Images: Detection, Alignment, and Recognition}},
  ADDRESS = {Marseille, France},
  ORGANIZATION = {{Erik Learned-Miller and Andras Ferencz and Fr{\'e}d{\'e}ric Jurie}},
  YEAR = {2008},
  MONTH = Oct,
  HAL_ID = {inria-00321923},
  HAL_VERSION = {v1},
}

@misc{dataset_taras_dog_breeds,
  author       = {AtharvaTaras},
  title        = {{Dog‑Breeds‑Dataset}: A dataset of images for dog breeds recognized by the FCI},
  howpublished = {\url{https://github.com/AtharvaTaras/Dog-Breeds-Dataset}},
  year         = {2025},
  note         = {Accessed: 2025‑12‑11; licensed under CC‑BY‑4.0},
  url          = {https://github.com/AtharvaTaras/Dog-Breeds-Dataset}
}

@misc{sparse_peft,
  author       = {Leonardo Santiago Benitez Pereira},
  title        = {sparse-peft},
  howpublished = {\url{https://github.com/LeonardoSanBenitez/sparse-peft}},
  note         = {Accessed: 2025-11-29},
  year         = {2025}
}

@InProceedings{George2025,
    author    = {George, Naveen and Dasaraju, Karthik Nandan and Chittepu, Rutheesh Reddy and Mopuri, Konda Reddy},
    title     = {The Illusion of Unlearning: The Unstable Nature of Machine Unlearning in Text-to-Image Diffusion Models},
    booktitle = CVPR,
    month     = {June},
    year      = {2025},
    pages     = {13393-13402}
}

@InProceedings{Zhou2025Delete,
    author    = {Zhou, Yu and Zheng, Dian and Mo, Qijie and Lu, Renjie and Lin, Kun-Yu and Zheng, Wei-Shi},
    title     = {Decoupled Distillation to Erase: A General Unlearning Method for Any Class-centric Tasks},
    booktitle = CVPR,
    month     = {June},
    year      = {2025},
    pages     = {20350-20359}
}

@misc{Quan2025Purge,
      title={Efficient Verified Machine Unlearning For Distillation}, 
      author={Yijun Quan and Zushu Li and Giovanni Montana},
      year={2025},
      eprint={2503.22539},
      archivePrefix={arXiv},
      primaryClass={cs.LG},
      url={https://arxiv.org/abs/2503.22539}, 
}

@misc{Lee2025UNDO,
      title={Distillation Robustifies Unlearning}, 
      author={Bruce W. Lee and Addie Foote and Alex Infanger and Leni Shor and Harish Kamath and Jacob Goldman-Wetzler and Bryce Woodworth and Alex Cloud and Alexander Matt Turner},
      year={2025},
      eprint={2506.06278},
      archivePrefix={arXiv},
      primaryClass={cs.LG},
      url={https://arxiv.org/abs/2506.06278}, 
}

@inproceedings{Frankle2019,
  title     = {The Lottery Ticket Hypothesis: Finding Sparse, Trainable Neural Networks},
  author    = {Frankle, Jonathan and Carbin, Michael},
  booktitle = ICLR,
  year      = {2019},
  publisher = {OpenReview.net},
  url       = {https://openreview.net/forum?id=rJl-b3RcF7},
  note      = {Published conference version; see also arXiv:1803.03635}
}

@misc{Lee2025Location,
      title={Does Localization Inform Unlearning? A Rigorous Examination of Local Parameter Attribution for Knowledge Unlearning in Language Models}, 
      author={Hwiyeong Lee and Uiji Hwang and Hyelim Lim and Taeuk Kim},
      year={2025},
      eprint={2505.16252},
      archivePrefix={arXiv},
      primaryClass={cs.CL},
      url={https://arxiv.org/abs/2505.16252}, 
}

@INPROCEEDINGS{cao2015,
  author={Cao, Yinzhi and Yang, Junfeng},
  booktitle={2015 IEEE Symposium on Security and Privacy}, 
  title={Towards Making Systems Forget with Machine Unlearning}, 
  year={2015},
  volume={},
  number={},
  pages={463-480},
  doi={10.1109/SP.2015.35},
  ISSN={2375-1207},
  month={May},}

@article{balaji2022ediff,
  title={ediff-i: Text-to-image diffusion models with an ensemble of expert denoisers},
  author={Balaji, Yogesh and Nah, Seungjun and Huang, Xun and Vahdat, Arash and Song, Jiaming and Zhang, Qinsheng and Kreis, Karsten and Aittala, Miika and Aila, Timo and Laine, Samuli and others},
  journal={arXiv preprint arXiv:2211.01324},
  year={2022}
}

@article{hertz2022prompttoprompt,
  title={Prompt-to-prompt image editing with cross attention control},
  author={Hertz, Amir and Mokady, Ron and Tenenbaum, Jay and Aberman, Kfir and Pritch, Yael and Cohen-Or, Daniel},
  journal={arXiv preprint arXiv:2208.01626},
  year={2022}
}

@article{rombach2022stableDiffusion,
  title={High-Resolution Image Synthesis with Latent Diffusion Models},
  author={Robin Rombach and A. Blattmann and Dominik Lorenz and Patrick Esser and Bj{\"o}rn Ommer},
  year={2021},
  pages={10674-10685},
  url={https://api.semanticscholar.org/CorpusID:245335280},
  journal=CVPR
}

@inproceedings{ho2020denoising,
 author = {Ho, Jonathan and Jain, Ajay and Abbeel, Pieter},
 booktitle = NeurIPS,
 editor = {H. Larochelle and M. Ranzato and R. Hadsell and M.F. Balcan and H. Lin},
 pages = {6840--6851},
 publisher = {Curran Associates, Inc.},
 title = {Denoising Diffusion Probabilistic Models},
 url = {https://proceedings.neurips.cc/paper_files/paper/2020/file/4c5bcfec8584af0d967f1ab10179ca4b-Paper.pdf},
 volume = {33},
 year = {2020}
}

@inproceedings{Zhang2024UnlearnCanvas,
 author = {Zhang, Yihua and Fan, Chongyu and Zhang, Yimeng and Yao, Yuguang and Jia, Jinghan and Liu, Jiancheng and Zhang, Gaoyuan and Liu, Gaowen and Kompella, Ramana and Liu, Xiaoming and Liu, Sijia},
 booktitle = NeurIPS,
 doi = {10.52202/079017-3055},
 editor = {A. Globerson and L. Mackey and D. Belgrave and A. Fan and U. Paquet and J. Tomczak and C. Zhang},
 pages = {96387--96423},
 publisher = {Curran Associates, Inc.},
 title = {UnlearnCanvas:  Stylized Image Dataset for Enhanced Machine Unlearning Evaluation in Diffusion Models},
 url = {https://proceedings.neurips.cc/paper_files/paper/2024/file/aebf4822d30c3f2600566af7eba83548-Paper-Datasets_and_Benchmarks_Track.pdf},
 volume = {37},
 year = {2024}
}

@inproceedings{Cai2025Slug,
  author    = {Zikui Cai and Yaoteng Tan and M. Salman Asif},
  title     = {Targeted Unlearning with Single Layer Unlearning Gradient},
  booktitle = ICML,
  year      = {2025},
  url       = {https://openreview.net/forum?id=6Ofb0cGXb5}
}

@inproceedings{Cywinski2025,
  title={SAeUron: Interpretable Concept Unlearning in Diffusion Models with Sparse Autoencoders},
  author={Cywi{\'n}ski, Bartosz and Deja, Kamil},
  booktitle=ICML,
  year={2025}
}

@InProceedings{Wu2024_scissorhands,
    author="Wu, Jing
    and Harandi, Mehrtash",
    editor="Leonardis, Ale{\v{s}}
    and Ricci, Elisa
    and Roth, Stefan
    and Russakovsky, Olga
    and Sattler, Torsten
    and Varol, G{\"u}l",
    title="Scissorhands: Scrub Data Influence via Connection Sensitivity in Networks",
    booktitle=ECCV,
    year="2025",
    publisher="Springer Nature Switzerland",
    address="Cham",
    pages="367--384",
    isbn="978-3-031-72970-6"
}

@misc{
    Wu2024EraseDiff,
    title={EraseDiff: Erasing Data Influence in Diffusion Models},
    author={Jing Wu and Trung Le and Munawar Hayat and Mehrtash Harandi},
    year={2024},
    url={https://openreview.net/forum?id=eVpjeCNsR6}
}

@inproceedings{Lyu2023SPM,
    author = {Lyu, Mengyao and Yang, Yuhong and Hong, Haiwen and Chen, Hui and Jin, Xuan and He, Yuan and Xue, Hui and Han, Jungong and Ding, Guiguang},
    year = {2024},
    month = {06},
    pages = {7559-7568},
    title = {One-dimensional Adapter to Rule Them All: Concepts, Diffusion Models and Erasing Applications},
    doi = {10.1109/CVPR52733.2024.00722},
    booktitle = CVPR
}

@inproceedings{li2024SEOT,
title={Get What You Want, Not What You Don't: Image Content Suppression for Text-to-Image Diffusion Models},
author={Senmao Li and Joost van de Weijer and taihang Hu and Fahad Khan and Qibin Hou and Yaxing Wang and jian Yang},
booktitle=ICLR,
year={2024},
url={https://openreview.net/forum?id=zpVPhvVKXk}
}

@inproceedings{kumari2023,
  author = {Kumari, Nupur and Zhang, Bingliang and Wang, Sheng-Yu and Shechtman, Eli and Zhang, Richard and Zhu, Jun-Yan},
  title = {Ablating Concepts in Text-to-Image Diffusion Models},
  booktitle = ICCV,
  year = {2023}
}

@inproceedings{Gandikota2023UCE,
  title={Unified Concept Editing in Diffusion Models},
  author={Gandikota, Rohit and Orgad, Hadas and Belinkov, Yonatan and Materzy{\'n}ska, Joanna and Bau, David},
  booktitle={Proceedings of the IEEE/CVF Winter Conference on Applications of Computer Vision},
  year={2024},
  note={arXiv:2308.14761}
}

@article{gandikota2023,
  title={Erasing Concepts from Diffusion Models},
  author={Rohit Gandikota and Joanna Materzynska and Jaden Fiotto-Kaufman and David Bau},
  journal=ICCV,
  year={2023},
  pages={2426-2436},
  url={https://api.semanticscholar.org/CorpusID:257495777}
}

@InProceedings{Zhang2023_fmn,
    author    = {Zhang, Gong and Wang, Kai and Xu, Xingqian and Wang, Zhangyang and Shi, Humphrey},
    title     = {Forget-Me-Not: Learning to Forget in Text-to-Image Diffusion Models},
    booktitle = CVPRw,
    month     = {June},
    year      = {2024},
    pages     = {1755-1764}
}

@misc{wang2024selectiveforgetting,
      title={Selective Forgetting: Advancing Machine Unlearning Techniques and Evaluation in Language Models}, 
      author={Lingzhi Wang and Xingshan Zeng and Jinsong Guo and Kam-Fai Wong and Georg Gottlob},
      year={2024},
      eprint={2402.05813},
      archivePrefix={arXiv},
      primaryClass={cs.CL},
      url={https://arxiv.org/abs/2402.05813}, 
}

@misc{chen2025scoreforgettingdistillation,
      title={Score Forgetting Distillation: A Swift, Data-Free Method for Machine Unlearning in Disffusion Models}, 
      author={Tianqi Chen and Shujian Zhang and Mingyuan Zhou},
      year={2025},
      eprint={2409.11219},
      archivePrefix={arXiv},
      primaryClass={cs.CV},
      url={https://arxiv.org/abs/2409.11219}, 
}

@inproceedings{golatkar2020,
  author={Golatkar, Aditya and Achille, Alessandro and Soatto, Stefano},
  booktitle=CVPR, 
  title={Eternal Sunshine of the Spotless Net: Selective Forgetting in Deep Networks}, 
  year={2020},
  volume={},
  number={},
  pages={9301-9309},
  doi={10.1109/CVPR42600.2020.00932},
  ISSN={2575-7075},
  month={June}}

@ARTICLE{shaik2024,
  author={Shaik, Thanveer and Tao, Xiaohui and Xie, Haoran and Li, Lin and Zhu, Xiaofeng and Li, Qing},
  journal={IEEE Transactions on Neural Networks and Learning Systems}, 
  title={Exploring the Landscape of Machine Unlearning: A Comprehensive Survey and Taxonomy}, 
  year={2024},
  volume={},
  number={},
  pages={1-21},
  doi={10.1109/TNNLS.2024.3486109},
  ISSN={2162-2388},
  month={},
}

@article{fan2023salun,
  title={Salun: Empowering machine unlearning via gradient-based weight saliency in both image classification and generation},
  author={Fan, Chongyu and Liu, Jiancheng and Zhang, Yihua and Wei, Dennis and Wong, Eric and Liu, Sijia},
  journal={arXiv preprint arXiv:2310.12508},
  year={2023}
}

@article{izzo2020,
  author       = {Zachary Izzo and
                  Mary Anne Smart and
                  Kamalika Chaudhuri and
                  James Y. Zou},
  title        = {Approximate Data Deletion from Machine Learning Models: Algorithms
                  and Evaluations},
  journal      = {CoRR},
  volume       = {abs/2002.10077},
  year         = {2020},
  url          = {https://arxiv.org/abs/2002.10077},
  eprinttype    = {arXiv},
  eprint       = {2002.10077},
  bibsource    = {dblp computer science bibliography, https://dblp.org}
}

@misc{zhang2023,
      title={Composing Parameter-Efficient Modules with Arithmetic Operations}, 
      author={Jinghan Zhang and Shiqi Chen and Junteng Liu and Junxian He},
      year={2023},
      eprint={2306.14870},
      archivePrefix={arXiv},
      primaryClass={cs.CL},
      url={https://arxiv.org/abs/2306.14870}, 
}

@misc{hu2021lora,
      title={LoRA: Low-Rank Adaptation of Large Language Models}, 
      author={Edward J. Hu and Yelong Shen and Phillip Wallis and Zeyuan Allen-Zhu and Yuanzhi Li and Shean Wang and Lu Wang and Weizhu Chen},
      year={2021},
      eprint={2106.09685},
      archivePrefix={arXiv},
      primaryClass={cs.CL},
      url={https://arxiv.org/abs/2106.09685}, 
}

@InProceedings{Radford2021Clip,
  title = 	 {Learning Transferable Visual Models From Natural Language Supervision},
  author =       {Radford, Alec and Kim, Jong Wook and Hallacy, Chris and Ramesh, Aditya and Goh, Gabriel and Agarwal, Sandhini and Sastry, Girish and Askell, Amanda and Mishkin, Pamela and Clark, Jack and Krueger, Gretchen and Sutskever, Ilya},
  booktitle = 	 ICML,
  pages = 	 {8748--8763},
  year = 	 {2021},
  editor = 	 {Meila, Marina and Zhang, Tong},
  volume = 	 {139},
  series = 	 {Proceedings of Machine Learning Research},
  month = 	 {18--24 Jul},
  publisher =    {PMLR},
  url = 	 {https://proceedings.mlr.press/v139/radford21a.html}
}

@misc{heusel2018fidmetric,
      title={GANs Trained by a Two Time-Scale Update Rule Converge to a Local Nash Equilibrium}, 
      author={Martin Heusel and Hubert Ramsauer and Thomas Unterthiner and Bernhard Nessler and Sepp Hochreiter},
      year={2018},
      eprint={1706.08500},
      archivePrefix={arXiv},
      primaryClass={cs.LG},
      url={https://arxiv.org/abs/1706.08500}, 
}

@misc{xu2023,
      title={Parameter-Efficient Fine-Tuning Methods for Pretrained Language Models: A Critical Review and Assessment}, 
      author={Lingling Xu and Haoran Xie and Si-Zhao Joe Qin and Xiaohui Tao and Fu Lee Wang},
      year={2023},
      eprint={2312.12148},
      archivePrefix={arXiv},
      primaryClass={cs.CL},
      url={https://arxiv.org/abs/2312.12148}, 
}

@misc{martens2015kronecker,
      title={Optimizing Neural Networks with Kronecker-factored Approximate Curvature}, 
      author={James Martens and Roger Grosse},
      year={2015},
      eprint={1503.05671},
      archivePrefix={arXiv},
      primaryClass={cs.LG},
      url={https://arxiv.org/abs/1503.05671}, 
}

@misc{clavera2024,
  title        = {Machine Unlearning: Fisher Information Matrix and Selective Forgetting in Deep Networks},
  author       = {Lluc Clavera Comas},
  year         = {2024},
  school       = {Universitat Politècnica de Catalunya (UPC) - BarcelonaTech},
  note         = {Bachelor's thesis, Facultat d'Informàtica de Barcelona (FIB)},
  url          = {https://upcommons.upc.edu/bitstream/handle/2117/410877/188878.pdf},
  month        = {June},
  day          = {25}
}

@inproceedings{munoz2024_sparse_peft,
    title = "{SQFT}: Low-cost Model Adaptation in Low-precision Sparse Foundation Models",
    author = "Munoz, Juan Pablo  and
      Yuan, Jinjie  and
      Jain, Nilesh",
    editor = "Al-Onaizan, Yaser  and
      Bansal, Mohit  and
      Chen, Yun-Nung",
    booktitle = "Findings of the Association for Computational Linguistics: EMNLP 2024",
    month = nov,
    year = "2024",
    address = "Miami, Florida, USA",
    publisher = "Association for Computational Linguistics",
    url = "https://aclanthology.org/2024.findings-emnlp.749",
    pages = "12817--12832",
}

@misc{choi2025mu_interpretation,
      title={Unlearning-based Neural Interpretations}, 
      author={Ching Lam Choi and Alexandre Duplessis and Serge Belongie},
      year={2025},
      eprint={2410.08069},
      archivePrefix={arXiv},
      primaryClass={cs.LG},
      url={https://arxiv.org/abs/2410.08069}, 
}

@article{Hinton2015self_distillation,
  title={Distilling the Knowledge in a Neural Network},
  author={Geoffrey E. Hinton and Oriol Vinyals and Jeffrey Dean},
  journal={ArXiv},
  year={2015},
  volume={abs/1503.02531},
  url={https://api.semanticscholar.org/CorpusID:7200347}
}
